%% file: main.tex
\documentclass[journal]{IEEEtran}
\usepackage{amsmath,amsfonts}
\usepackage{algorithmic}
\usepackage{algorithm}
\usepackage{array}
\usepackage[caption=false,font=normalsize,labelfont=sf,textfont=sf]{subfig}
\usepackage{textcomp}
\usepackage{stfloats}
\usepackage{url}
\usepackage{verbatim}
\usepackage{graphicx}
\usepackage{cite}

\usepackage{graphicx}
\usepackage{algorithm}
\usepackage{algorithmic}

\usepackage{newfloat}
\usepackage{listings}
\usepackage{amsmath}
\usepackage{amssymb}
\usepackage{graphicx}
\usepackage[table,xcdraw]{xcolor}
\usepackage{multirow}

\usepackage{bm}
\DeclareCaptionStyle{ruled}{labelfont=normalfont,labelsep=colon,strut=off}
\floatstyle{ruled}

\usepackage{amsmath,amsfonts,bm}

\usepackage{xcolor}
\usepackage{pifont}
\newif\ifshowrevisions
\showrevisionsfalse

\newcommand{\revised}[1]{%
  \ifshowrevisions
    \textcolor{red}{#1}%
  \else
    #1%
  \fi
}

\newcommand{\revisedblock}{%
  \ifshowrevisions
    \color{red}%
  \fi
}

\begin{document}

\title{A Gray-box Attack against Latent Diffusion Model-based Image Editing by Posterior Collapse}

\author{Zhongliang Guo,
    Chun Tong Lei,
    Lei Fang,
    Shuai Zhao,
    Yifei Qian,
    Jingyu Lin,
    Zeyu Wang,\\
    Cunjian Chen,, 
    Ognjen Arandjelovi\'c,
    Chun Pong Lau


\thanks{Zhongliang Guo, Lei Fang, and Ognjen Arandjelovi\'c are with the School of Computer Science, University of St Andrews, United Kingdom (E-mail: \{zg34, lf28, oa7\}@st-andrews.ac.uk).}
\thanks{Chun Tong Lei and Chun Pong Lau are with the Department of Data Science, City University of Hong Kong, China (E-mail: \{ctlei2, cplau27\}@cityu.edu.uk).}
\thanks{Shuai Zhao is with the College of Computing and Data Science, Nanyang Technological University, Singapore (E-mail: shuai.zhao@ntu.edu.sg).}
\thanks{Yifei Qian is with the School of Computer Science, University of Nottingham, United Kingdom (E-mail: yifei.qian@nottingham.ac.uk).}
\thanks{Jinyu Lin and Cunjian Chen are with the 	
Department of Data Science and Artificial Intelligence, Monash University, Australia (E-mail: \{jingyu.lin, cunjian.chen\}@monash.edu).}
\thanks{Zeyu Wang is with the 	
College of Information Science \& Electronic Engineering, Zhejiang University, Hangzhou, China (E-mail: wangzeyu2020@zju.edu.cn).}
\thanks{Corresponding Author(s): Chun Pong Lau and Shuai Zhao.}
}

\markboth{Arxiv Preprint}%
{Shell \MakeLowercase{\textit{et al.}}: A Sample Article Using IEEEtran.cls for IEEE Journals}


\maketitle

\begin{abstract}
Recent advancements in Latent Diffusion Models (LDMs) have revolutionized image synthesis and manipulation, raising significant concerns about data misappropriation and intellectual property infringement. While adversarial attacks have been extensively explored as a protective measure against such misuse of generative AI, current approaches are severely limited by their heavy reliance on model-specific knowledge and substantial computational costs.
Drawing inspiration from the posterior collapse phenomenon observed in VAE training, we propose the Posterior Collapse Attack (PCA), a novel framework for protecting images from unauthorized manipulation. Through comprehensive theoretical analysis and empirical validation, we identify two distinct collapse phenomena during VAE inference: diffusion collapse and concentration collapse. Based on this discovery, we design a unified loss function that can flexibly achieve both types of collapse through parameter adjustment, each corresponding to different protection objectives in preventing image manipulation.
\revised{Our method significantly reduces dependence on model-specific knowledge by requiring access to only the VAE encoder, which constitutes less than 4\% of LDM parameters. Notably, PCA achieves prompt-invariant protection by operating on the VAE encoder before text conditioning occurs, eliminating the need for empty prompt optimization required by existing methods.} This minimal requirement enables PCA to maintain adequate transferability across various VAE-based LDM architectures while effectively preventing unauthorized image editing.
Extensive experiments show PCA outperforms existing techniques in protection effectiveness, computational efficiency (runtime and VRAM), and generalization across VAE-based LDM variants. Our code is available at \url{https://github.com/ZhongliangGuo/PosteriorCollapseAttack}.
\end{abstract}

\begin{IEEEkeywords}
Adversarial attack, diffusion model, imperceptible attack, transferable attack.
\end{IEEEkeywords}

\section{Introduction}

\IEEEPARstart{T}{he} field of generative artificial intelligence has witnessed unprecedented advancements, particularly in the domain of image synthesis and manipulation. State-of-the-art methodologies, exemplified by diffusion models such as Stable Diffusion~\cite{ldm}, have demonstrated remarkable image generation quality and speed compared to other generation paradigm~\cite{li2024autoregressive}. These breakthroughs have revolutionized creative industries and democratized access to sophisticated image editing tools, empowering users across various domains.

However, the proliferation of such powerful technologies inevitably engenders a concomitant set of ethical quandaries and potential security vulnerabilities. Of particular concern is the risk of data misappropriation and intellectual property infringement. The facility with which diffusion-based models can be employed to manipulate existing visual content presents a significant challenge to the integrity of digital assets. For instance, malicious actors could exploit Stable Diffusion to edit copyrighted images, effectively ``laundering'' them by removing or altering distinctive features~\cite{ldm}. This process of ``de-copyrighting'' not only undermines the rights of content creators but also poses a threat to the economic ecosystems built around digital imagery, potentially destabilizing industries ranging from photography to digital art.

In the realm of machine learning security, adversarial attacks have emerged as a critical area of study~\cite{tifs1}. These attacks aim to perturb the input of models in ways that are imperceptible to humans but can significantly disrupt the  output.
This concept of adversarial attacks provides a promising solution to the misuse of generative AI~\cite{liu2025diffprotect}. By introducing carefully crafted, acceptable perturbations to data that could be misused, it may be possible to impede the ability of diffusion models to edit or manipulate such content effectively.

Current protection techniques can be broadly categorized based on their objectives: some aim to preserve the original image by minimizing perturbation distance, while others maximize deviation from the expected edited output. However, these approaches often rely heavily on extensive prior knowledge, specifically requiring full white-box access to the parameters of the target models~\cite{advdm,photoguard,mist,sds}.
Specifically, existing methods~\cite{advdm,photoguard,mist,sds} were naturally designed to target the U-Net, the primary generative component of LDMs, with their loss computations fundamentally requiring U-Net participation. While this focus on the core generation module is intuitive, it necessitates access much more model parameters and introduces substantial computational overhead.
This assumption is largely impractical in real-world scenarios, especially given the rapid pace of technological advancement in generative AI. The proliferation of model architectures and continuous evolution of backbone networks create a dynamic landscape in which protection methods quickly become obsolete. Developing a universal method that can effectively protect against the vast majority of generative models without detailed knowledge of their internal structures is exceedingly difficult.
Furthermore, the diversity of model architectures means that a method designed for one particular model or family may be entirely ineffective against others. This lack of generalizability severely limits the practical applicability of such approaches in a rapidly changing technological environment.

\begin{figure*}[tb]
    \centering
    {\revisedblock
    \includegraphics[width=\linewidth]{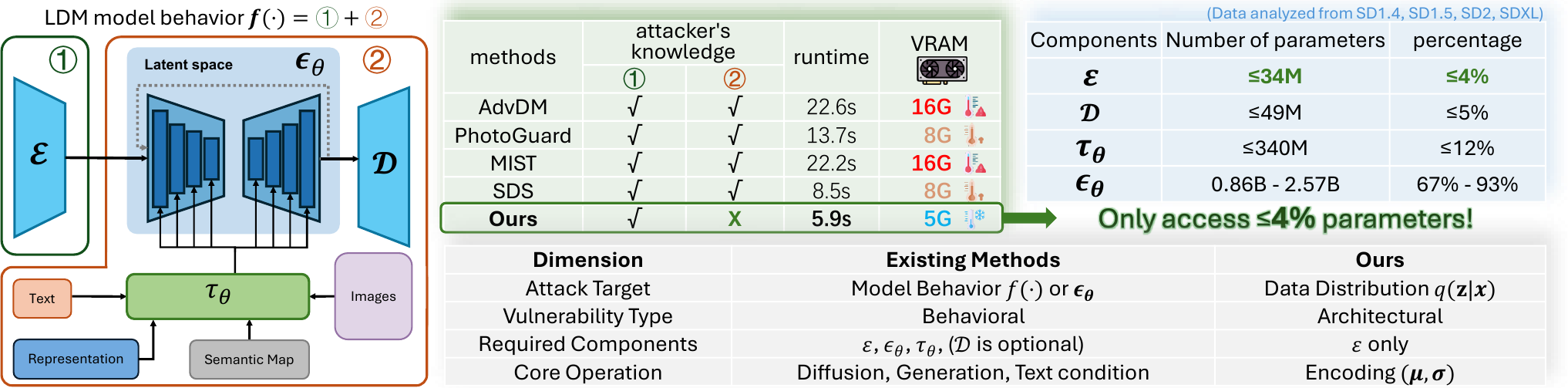}
    \caption{Comparison of our proposed method with AdvDM~\cite{advdm}, PhotoGuard~\cite{photoguard}, MIST~\cite{mist}, and SDS~\cite{sds}. Our method only requires access to the encoder of the VAE, achieving equal or superior performance on multi scenarios (Fig.~\ref{fig:demo}) with comparable image editing semantic degrade, lower runtime, and less occupied VRAM. All runtime and VRAM measurements were conducted on a NVIDIA RTX 3090 with $T=40$ iterations and 512×512 resolution.}
    \label{fig:teaser}}
    \vspace{1em}
    
    \includegraphics[width=\linewidth]{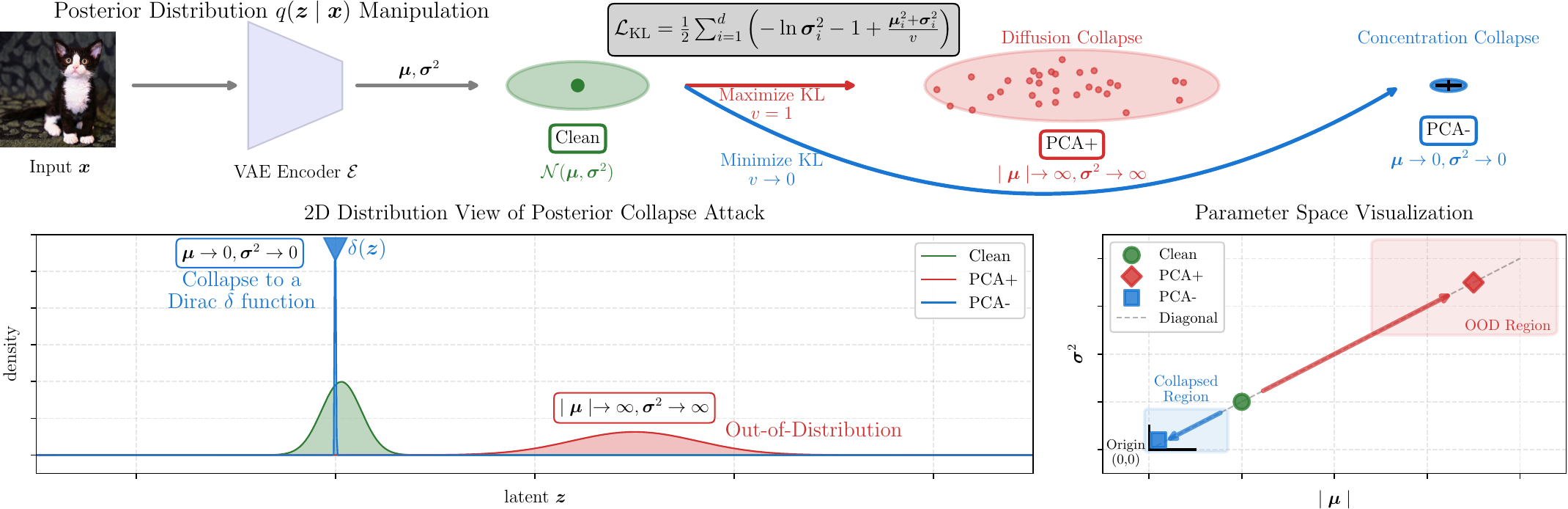}
    \caption{Overview of the Posterior Collapse Attack (PCA). The VAE encoder maps input images to posterior distributions $q(\boldsymbol{z}|\boldsymbol{x})=\mathcal{N}(\boldsymbol{\mu}, \boldsymbol{\sigma}^2)$. By manipulating the KL divergence with different hyperparameter $v$, PCA induces two types of collapse: \textbf{PCA+} (Objective 1) maximizes KL to create diffusion collapse, pushing the distribution out-of-distribution with $|\boldsymbol{\mu}| \to \infty$ and $\boldsymbol{\sigma}^2 \to \infty$, thereby preserving original image content; \textbf{PCA-} (Objective 2) minimizes KL to create concentration collapse, collapsing the distribution to a Dirac $\delta$ function with $\boldsymbol{\mu} \to 0$ and $\boldsymbol{\sigma}^2 \to 0$, thereby destroying semantic information.}
\label{fig:framework}
\end{figure*}
Given these limitations in current protective measures and the narrow scope of existing attack methods on diffusion models, addressing these critical gaps requires a solution that satisfies several key criteria:
\begin{itemize}
    \item To ensure broader applicability and resilience against rapid technological advancements, the method should minimize dependence on white-box information of the target models or merely rely on white-box information that is public across the most popular current methods.
    \item The protection mechanism should maintain high efficiency while requiring minimal computational overhead, making it practical for scenarios where computational resources may be constrained.
    \item For the effectiveness, the method should exhibit better transferability, easily generalizing across various models and architectures.
\end{itemize}

In light of these challenges, we aim to propose a new protection method targeting Variational Autoencoders (VAEs) hinge on a key observation: the vast majority of state-of-the-art generative models leveraging VAEs, where VAEs perform as a upstream bottleneck.
This commonality presents a promising opportunity to design a more universal attack strategy targeting the LDMs, potentially addressing the need for a generalizable protection method.

In literature, existing attacks on VAEs primarily focus on corrupting the decoding process, leading to deteriorated reconstruction~\cite{kuzina2022alleviating}, which does not align with our objective of disrupting the downstream diffusion process.
Notably, the encoding posterior distribution of VAEs, often overlooked in previous attack methods, provides a more comprehensive representation of the latent space, which is the bottleneck of all downstream tasks,
potentially leading to more effective and generalizable attacks.

Drawing inspiration from the phenomenon of posterior collapse in VAEs' \textbf{training}~\cite{razavi2018preventing}, in which the latent variables fail to capture meaningful information from the input data, we propose the Posterior Collapse Attack (PCA) to intentionally induce this collapse-like phenomenon in the \textbf{inference} stage. As shown in Fig.~\ref{fig:framework}, our exploration reveals two distinct types of collapse during inference: diffusion collapse and concentration collapse, each corresponding to a different protection objective (preserving image content versus disrupting edits). Through our proposed novel loss function, PCA can flexibly achieve either protection objective by inducing the corresponding type of collapse through parameter adjustment.

\begin{figure*}[t]
    \centering
    \includegraphics[width=\textwidth]{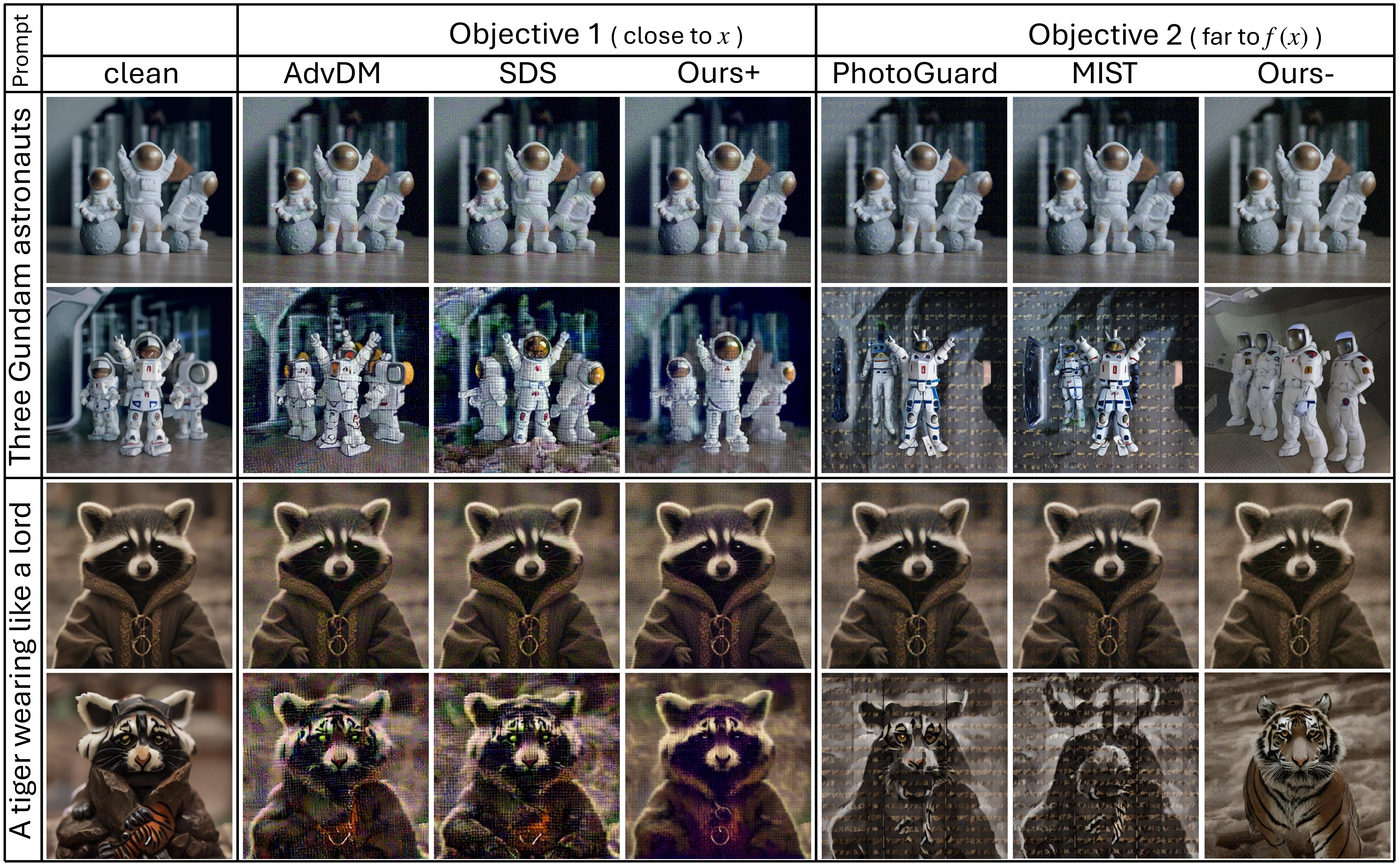}
    \caption{Comparison of our method with other baselines subject to different Objectives.
    The first column shows editing prompts.
    The second column shows the original input images and expected edited outputs.
    The subsequent columns show images protected by different methods and their corresponding outputs. Notably, $x$ refers to the original image will be edited by LDM; $f(\cdot)$ refers to the LDM-based image editing.}
    \label{fig:demo}
\end{figure*}

As illustrated in Fig.~\ref{fig:teaser} and Fig.~\ref{fig:demo}, our method effectively prevents unauthorized image manipulation in a near black-box manner (less than 4\% parameters).
This approach offers a more robust and transferable solution.
Unlike previous methods that rely heavily on model-specific knowledge, our method merely leverages encoder of VAEs, providing a more universally applicable protection mechanism against unauthorized image manipulation across various generative AI architectures.

Our contributions can be summarized as follows:
\begin{itemize}
    \item We characterize two distinct collapse phenomena in VAE inference, diffusion collapse and concentration collapse, establishing their connection to protection objectives and bridging collapse mechanisms with protection strategies.
    \item We design a unified loss function achieving both protection types through parameter adjustment, eliminating separate methods for different goals.
    \item Extensive experiments demonstrate that PCA achieves strong protection performance while requiring minimal white-box information, and exhibits better transferability across various diffusion models and architectures.
\end{itemize}

\section{Related Work}

\subsection{Generation Models}
Diffusion Probabilistic Model (DPM)~\cite{ho2020ddpm} has achieved state-of-the-art results in density estimation~\cite{kingma2021variational} and sample quality~\cite{dhariwal2021diffusionbeatgan}. These models are powerful due to their U-Net backbone, which suits the inductive biases of image-like data. However, they face challenges in inference speed and training costs, especially for high-resolution images. To address these limitations, researchers have developed approaches like Denoising Diffusion Implicit Models (DDIM)~\cite{song2020ddim} to enhance sampling speeds, and explored two-stage processes. Latent Diffusion Models (LDM)~\cite{ldm} use autoencoding models to learn a perceptually similar space with lower computational complexity, while VQ-VAEs~\cite{yan2021videogpt,razavi2019vqvae2} and VQGANs~\cite{yu2021vectorvqgan,vqgan} utilize discretized latent spaces and adversarial objectives to scale to larger images. These advancements aim to overcome the challenges of complex training, data shortages, and computational costs associated with diffusion models, particularly for high-resolution image synthesis.

\subsection{Adversarial Attack}

The field of adversarial machine learning was catalyzed by the seminal work of Szegedy et al.~\cite{szegedy2013intriguing}, who first uncovered the vulnerability of neural networks.
Subsequent research has led to the development of various attack and defense methodologies~\cite{goodfellow2014explaining,madry2017towards,tifs3}.
The implications of adversarial attacks raised significant concerns in critical applications. This has spurred the development of defensive strategies, such as adversarial training ~\cite{lau2023interpolated,tifs2} and input transformation~\cite{lin2020dual}.
We have seen the expansion of adversarial attack research into more complex domains~\cite{liu2023instruct2attack,guo2024white,li2024threats,zhao2024survey}.

\subsubsection{Attack on VAEs}
Early works of adversarial attacks on VAEs primarily focused on attacking the image reconstruction capabilities of VAEs. Various attack methods against VAE and VAE-GAN models were developed~\cite{tabacof2016adversarial,gondim2018adversarial,kos2018adversarial}, including classifier-based attacks, attacks using the VAE loss function, and latent space attacks. These methods aimed to generate adversarial examples that would be reconstructed as images from different classes. Recent research has expanded to investigating attacks on the encoder output of hierarchical VAEs, considering the higher-level latent variables~\cite{kuzina2022alleviating}.
Most existing research on VAE attacks has focused on manipulating the input or latent space to affect reconstruction output.
Our approach, however, differs significantly from these previous works. We specifically target the disruption of downstream tasks based on VAE encodings, such as diffusion models, rather than the VAEs' own reconstruction. Furthermore, we attack the distributional parameters ($\boldsymbol{\mu},\boldsymbol{\sigma}^2$) defining the approximate posterior $q(\boldsymbol{z}|\boldsymbol{x})$, rather than the sampled latent variables themselves. This allows us to explore vulnerabilities in VAE-based systems where distributional properties are critical for downstream processing.

\subsubsection{Attack on Diffusion-based Image Editing}
\revised{Recent research has paid attention to adversarial attacks against diffusion models for image editing. \cite{advdm} targeted the noise prediction module, while \cite{photoguard} and \cite{shan2023glaze} minimized latent space distances. \cite{mist} combined semantic and textural losses for protection. \cite{sds} revealed the encoder's vulnerability in latent diffusion models and proposed using Score Distillation Sampling to reduce computational costs.
However, as shown in Fig.~\ref{fig:teaser}, these approaches often rely heavily on full white-box access to target models or extensive knowledge of fundamental generative model principles. Specifically, the design of existing methods fundamentally requiring U-Net participation alongside Text Encoder and (optionally) VAE Decoder. They exploit behavioral vulnerabilities in generation process, which cannot be trivially adapted to methods using only VAE encoder that targets architectural vulnerabilities in posterior distribution.
This dependency limits their practical applicability, especially given the rapid evolution of generative AI architectures.
There is a need for methods that are less dependent on model-specific knowledge, can significantly degrade generation quality, and exhibit better transferability across various models.}

\section{Preliminary}
Perceptual compression models typically use an autoencoder trained with a perceptual loss \cite{zhang2018unreasonable} and a patch-based \cite{isola2017image} adversarial objective \cite{dosovitskiy2016generating}. Specifically, given an image $\boldsymbol{x}\in\mathbb{R}^{H\times W\times3}$ in pixel space, $\boldsymbol{x}$ is encoded into latent representation $z = \mathcal{E}(\boldsymbol{x}) \in \mathbb{R}^d$ by an autoencoder. A decoder $\mathcal{D}$ reconstructs the image from the latent, giving $\tilde{\boldsymbol{x}} = \mathcal{D}(\boldsymbol{z}) = \mathcal{D}(\mathcal{E}(\boldsymbol{x}))$, where $\boldsymbol{x}\in\mathbb{R}^{H\times W\times 3}$. The encoder downsamples the image by a factor $\mathit{f} = H/h = W/w$, $\mathit{f}$ is often set as 8 in latent diffusion models.

We focus on VAE, which is used for most the state-of-the-art LDMs.
VAE can be considered as a probabilistic generative extension of the ordinary autoencoders.
The encoder of a VAE aims at approximating the intractable posterior distribution of the latent variable $\boldsymbol{z}$ by a Gaussian distribution $q(\boldsymbol{z}|\boldsymbol{x})=N(\boldsymbol{\mu}, \text{diag}(\boldsymbol{\sigma}^2))$.
The training proceeds by minimising the KL divergence between the approximating variational distribution $q(\boldsymbol{z})$ and the true posterior $p(\boldsymbol{z}|\boldsymbol{x})$.

The Latent Diffusion Model is one of the most popular image generation or editing methods. During training, an image $\boldsymbol{x}_0$ and a condition $\boldsymbol{c}$ are used as inputs. The image is encoded into a latent code $\boldsymbol{z}_0$ by an image encoder, such as VAE or its variants. This latent code $\boldsymbol{z}_0$ is then progressively combined with Gaussian noise $\epsilon$ through a forward process, which can be summarized as follows:
\begin{equation}
    \boldsymbol{z}_{t_{\text{diff}}}=\sqrt{\bar{\alpha}_{t_{\text{diff}}}}\boldsymbol{z}_0+\sqrt{1-\bar{\alpha}_{t_{\text{diff}}}}\epsilon,\epsilon\sim\mathcal{N}(\mathbf{0},\mathbf{I}),
\end{equation}
where $\bar{\alpha}_{t_{\text{diff}}}=\prod_{i=1}^{t_{\text{diff}}}\alpha_{i},\alpha_{i}\in(0,1)$.

The diffusion model is trained to approximate the original data distribution
with a denoising objective:
\begin{equation}
    \mathbb{E}_{\boldsymbol{z}_0,\boldsymbol{c},\epsilon,{t_{\text{diff}}}}[\|\epsilon_\theta(\boldsymbol{z}_{t_{\text{diff}}},\boldsymbol{c},{t_{\text{diff}}})-\epsilon\|],
\end{equation}
where $\epsilon_\theta$ is the model prediction. During inference, with a random Gaussian noise start $\boldsymbol{z}_{T_{\text{diff}}}$ and condition $\boldsymbol{c}$, the diffusion model carries out the opposite process from step ${T_{\text{diff}}}$ down to 1 to get the encoding of sampled image $\hat{\boldsymbol{z}}_0$ by:
\begin{equation}
    \hat{\boldsymbol{z}}_{{t_{\text{diff}}}-1}=\frac{1}{\sqrt{\alpha_{t_{\text{diff}}}}}\left(\hat{\boldsymbol{z}}_{t_{\text{diff}}}-\frac{1-\alpha_{t_{\text{diff}}}}{\sqrt{1-\bar{\alpha}_{t_{\text{diff}}}}}\epsilon_\theta(\hat{\boldsymbol{z}}_{t_{\text{diff}}},\boldsymbol{c},{t_{\text{diff}}})\right)+\sigma_{t_{\text{diff}}}\epsilon,
\end{equation}
where $\sigma_{{t_{\text{diff}}}}=\frac{1-\bar{\alpha}_{{t_{\text{diff}}}-1}}{1-\bar{\alpha}_{{t_{\text{diff}}}}}\beta_{{t_{\text{diff}}}},\beta_{{t_{\text{diff}}}}=1-\alpha_{{t_{\text{diff}}}}$.

\section{Method}

\subsection{Problem Definition}
Adversarial attack aims to craft an imperceptible perturbation $\boldsymbol{\delta}$, added on the clean image $\boldsymbol{x}$ as the adversarial sample $\boldsymbol{x}_{adv}$, resulting in the wrong or disruptive output of machine learning models. The key concept of the adversarial attack against LDM-based image editing can be summarized as two objectives:
\begin{center}
    \textbf{Objective 1}: $ \min_{\boldsymbol{\delta}} d(f(\boldsymbol{x}+\boldsymbol{\delta}),\boldsymbol{x})\,\,\,\,\,\,\,\,\,\,\,\,\,\,\,
       s.t.\,
       \|\boldsymbol{\delta}\|_{p} \leq \epsilon,$\\
\textbf{Objective 2}: $ \max_{\boldsymbol{\delta}} d(f(\boldsymbol{x}+\boldsymbol{\delta}),f(\boldsymbol{x}))\quad
       s.t.\,
       \|\boldsymbol{\delta}\|_{p} \leq \epsilon,$
\end{center}

where $f(\cdot)$ is a kind of LDM-based image editing method; $d(\cdot)$ measures the perceptual distance between two inputs; $\| \cdot \|_p$ applies the $\ell_\infty$ norm constraint to maintain visual integrity of the adversarial sample. Intuitively, Objective 1 preserves original image content by keeping the edited adversarial sample similar to $\boldsymbol{x}$, while Objective 2 disrupts unauthorized editing by maximizing deviation from the expected output $f(\boldsymbol{x})$. Both objectives prevent infringer-desired manipulations, with Objective 1 causing some distance between $f(\boldsymbol{x}+\boldsymbol{\delta})$ and $\boldsymbol{x}$ as a side effect.

Existing methods in the literature typically address either Objective 1 or Objective 2. However, the solving algorithms employed by these methods often require extensive white-box information about the target model, particularly access to the neural backbone U-Net of the LDM. This heavy reliance on model-specific details limits their transferability and applicability across different LDM architectures, requiring more computing resource.

Our method proposes a solving algorithm that can achieve both Objective 1 and Objective 2, but takes a fundamentally different approach. Instead of relying on detailed knowledge of the entire LDM pipeline, we exploit the inherent characteristics of LDM-based editing by targeting the VAE component, which is common across various LDM architectures.
This strategy allows us to achieve the goal of maximizing the disparity between $f(\boldsymbol{x}_{adv})$ and $f(\boldsymbol{x})$ without requiring extensive access to model-specific information, particularly the compute-intensive and model-specific U-Net component.
By concentrating on the VAE, our approach aligns more closely with real-world scenarios where full model access may not be available, providing an efficient solution to prevent infringers from exploiting LDM-based image editing outputs.

\subsection{Posterior Collapse of Variational Autoencoder}\label{sec:vae-collapse}
A key observation driving our approach is the ubiquity of VAEs in the architecture of LDMs.
VAEs serve as a foundational component across different LDM implementations, often with only minor variations between models.
For instance, the architecture of VAE used in Stable Diffusion 2.0 is similar as that of Stable Diffusion 1.5. This commonality presents a strategic opportunity: by focusing on the vulnerability of VAE architecture, we can potentially affect a wide range of LDMs without requiring detailed knowledge of their specific downstream tasks.

VAE can be considered as a probabilistic generative extension of the ordinary autoencoders. The encoder of a VAE aims at approximating the intractable posterior distribution of the latent variable $\boldsymbol{z}$ by a Gaussian distribution $q(z|x)=\mathcal{N}(\boldsymbol{\mu}, \text{diag}(\boldsymbol{\sigma}^2))$, where $\text{diag}$ denotes a diagonal matrix formed with vector $\boldsymbol{\sigma}^2$ as the diagonal entries. The training proceeds by minimising the KL divergence between the approximating variational distribution $q(\boldsymbol{z})$ and the true posterior $p(\boldsymbol{z}|\boldsymbol{x})$, and the learning objective is:
\begin{equation}
    \mathcal{L}(\boldsymbol{x}) = -\mathbb{E}_{q(\boldsymbol{z}|\boldsymbol{x})}\left [\log p(\boldsymbol{x}, \boldsymbol{z}) - \ln q(\boldsymbol{z}|\boldsymbol{x}) \right ].
\end{equation} The above loss is also known as negative ELBO. 
By observing the generating process $p(\boldsymbol{x},\boldsymbol{z}) = p(\boldsymbol{z})p(\boldsymbol{x}|\boldsymbol{z})$, the loss can be written alternatively as:
\begin{equation}\label{eq:vaeloss}
    \mathcal{L}(\boldsymbol{x}) = -\mathbb{E}_{q(\boldsymbol{z}|\boldsymbol{x})}\left [\ln p(\boldsymbol{x}|\boldsymbol{z}) \right ] +D_{\text{KL}}(q(\boldsymbol{z}|\boldsymbol{x}) \parallel p(\boldsymbol{z})),
\end{equation}
where the first term is known as reconstruction error and the second term, the KL divergence between the variational distribution and the prior distribution, usually a standard normal distribution, that is $p(\boldsymbol{z}) = \mathcal{N}(\mathbf{0},\mathbf{I})$. The KL divergence has a regularisation effect such that the variational distribution is ``close'' to a standard spherical Gaussian distribution. 

By general consensus, \textbf{training} of VAE usually suffers an optimization issue called posterior collapsing~\cite{razavi2018preventing}, that is the KL divergence term of Equation~\ref{eq:vaeloss} dominates the overall loss such that all posteriors collapse to the uninformative prior and the reconstruction error or the likelihood term is ignored.
We draw inspiration from this vulnerability and aim to propose adversarial attacks to cause the posterior collapse during \textbf{inference stage}.

\subsection{Unified Loss Function for Two Collapse Phenomena}\label{sec:detailed-collapses}

Building upon the posterior collapse phenomenon discussed in Section~\ref{sec:vae-collapse}, we now present our unified approach to inducing both diffusion collapse and concentration collapse during VAE inference. Our method (shown in Fig.~\ref{fig:framework}) leverages a single KL divergence-based loss function that can achieve both protection objectives through strategic parameter adjustment.

Posterior-Collapse-phenomenon in VAEs' inference stage can manifest in two primary forms: diffusion collapse and concentration collapse. Both potential collapses could be induced in the inference stage, by maximizing and minimizing the KL divergence. Here we give definitions of two kind of collapses and explain when the phenomenon will occur. Let \(\mathcal{X}\) be the input space and \(\mathcal{Z} \subseteq \mathbb{R}^d\) be the latent space of a VAE. The posterior distribution in a VAE is denoted by \(q(\boldsymbol{z} \mid \boldsymbol{x})\), where \(\boldsymbol{x} \in \mathcal{X}\) and \(\boldsymbol{z} \in \mathcal{Z}\),
\begin{enumerate}

    \item \textbf{Diffusion Collapse:} (for objective 1)

    We witness diffusion collapse when the posterior distribution becomes overly dispersed:
    \begin{equation}
q(\boldsymbol{z} \mid \boldsymbol{x}) \approx \mathcal{U}(\mathcal{Z}) \quad \text{as} \quad \boldsymbol{\sigma}^2 \rightarrow \infty,
\end{equation}
where \(\mathcal{U}(\mathcal{Z})\) denotes the uniform distribution over \(\mathcal{Z}\). In this case, the encoder produces highly uncertain latent representations for all inputs. This can be done by maximizing the KL divergence between $\mathcal{N}(\mathbf{0}, \mathbf{I})$ (i.e., setting $v=1$ in the Equation~\ref{eq:final:loss}), which can be formulated as:
\begin{equation}
\mathcal{L}_{\mathrm{KL}}(\boldsymbol{x}) = \frac{1}{2} \sum_{i=1}^d \left(-\ln \boldsymbol{\sigma}_i^2 - 1 + \boldsymbol{\mu}_i^2 + \boldsymbol{\sigma}_i^2\right).
\label{pca+}
\end{equation}
    
    \item \textbf{Concentration Collapse:} (for objective 2)

    In this scenario, the posterior distribution becomes excessively concentrated, approaching a Dirac delta function:
    \begin{equation}
q(\boldsymbol{z} \mid \boldsymbol{x}) \rightarrow \delta(\boldsymbol{z} - \boldsymbol{\mu}_0) \quad \text{as} \quad \boldsymbol{\mu}, \boldsymbol{\sigma}^2 \rightarrow \mathbf{0},
\end{equation}
where \(\delta\) is the Dirac delta function, and \(\boldsymbol{\mu}_0\) is typically the mean of the prior distribution. This collapse results in the encoder mapping all inputs to nearly identical latent representations. This can be done by minimizing the KL divergence between $\mathcal{N}(\mathbf{0}, \text{diag}(c))$, where $c$ is some arbitrary small number (i.e., setting $v\rightarrow0$ in theEquation~\ref{eq:final:loss}), and the KL divergence can be formulated as:
\begin{equation}
   \mathcal{L}_{\mathrm{KL}}(\boldsymbol{x}) = \frac{1}{2}\sum_{i=1}^{d}\left(\ln\frac{c}{\boldsymbol{\sigma}_i^2} -1 + \frac{\boldsymbol{\mu}_i^2+\boldsymbol{\sigma}_i^2}{c}\right).
\label{pca-}
\end{equation}

\end{enumerate}
Both forms can be derived from the general case, we aim to leverage the divergence measure between $q(\boldsymbol{z}|\boldsymbol{x})$ and a target distribution $p^\ast(\boldsymbol{z})$. The objective is to generate $\boldsymbol{x}_{adv}$ by minimising the KL divergence between $q(\boldsymbol{z}|\boldsymbol{x})$ and $p^\ast(\boldsymbol{z})$.
The KL divergence of two multivariate Gaussian distributions is:
\begin{equation}
\label{eq:KL_general}
\begin{split}
    D_{\text{KL}}(\mathcal{N}_1 \parallel \mathcal{N}_2) 
    = &\frac{1}{2}\Big[\ln\frac{|\boldsymbol{\Sigma}_2|}{|\boldsymbol{\Sigma}_1|} - d + \text{tr} \{ \boldsymbol{\Sigma}_2^{-1}\boldsymbol{\Sigma}_1 \} \\
    &+ (\boldsymbol{\mu}_2 - \boldsymbol{\mu}_1)^T \boldsymbol{\Sigma}_2^{-1}(\boldsymbol{\mu}_2 - \boldsymbol{\mu}_1)\Big].
\end{split}
\end{equation}
\noindent where $\mathcal{N}_1 = \mathcal{N}(\bm{\mu}_1, \boldsymbol{\Sigma}_1), \mathcal{N}_2 = \mathcal{N}(\bm{\mu}_2, \boldsymbol{\Sigma}_2)$.

Given two multivariate Gaussian distribution $\mathcal{N}_1(\boldsymbol{\mu}_1,\boldsymbol{\Sigma}_1)$ and $\mathcal{N}_2(\boldsymbol{\mu}_2,\boldsymbol{\Sigma}_2)$, where $\mathcal{N}_1$ and $\mathcal{N}_2$ refer to the posterior distribution $q(\boldsymbol{z}|\boldsymbol{x})$ and our target prior distribution $p^\ast(\boldsymbol{z})$.
We set the target attack $p^\ast(\boldsymbol{z})$ to a zero mean Gaussian distribution $p^\ast(\boldsymbol{z})=\mathcal{N}(\mathbf{0},v\mathbf{I})$, where $v>0$ is hyper-parameter that controls the disruptive effect. Since both $p^\ast(\boldsymbol{z})$ and $q(\boldsymbol{z}|\boldsymbol{x})$ are diagonal multivariate Gaussian distributions, Equation~\ref{eq:KL_general} can be derived as:
\begin{equation}
\begin{aligned}
    D_{\text{KL}}(\mathcal{N}_1\parallel\mathcal{N}_2) 
    &=\frac{1}{2}\left(\ln{\prod_{i=1}^{d} \frac{v}{\boldsymbol{\sigma}_i^2}}-d + \sum_{i=1}^{d}\frac{\boldsymbol{\sigma}_i^2}{v} + \frac{\boldsymbol{\mu}^T\boldsymbol{\mu}}{v}\right)\\
    & = \frac{1}{2}\left(\sum_{i=1}^{d}\ln{\frac{v}{\boldsymbol{\sigma}_i^2}}-d + \sum_{i=1}^{d}\frac{\boldsymbol{\sigma}_i^2}{v} + \sum_{i=1}^{d}\frac{\boldsymbol{\mu}_i^2}{v}\right)\\
    & = \frac{1}{2}\sum_{i=1}^{d}\left(\ln\frac{v}{\boldsymbol{\sigma}_i^2} -1 + \frac{\boldsymbol{\mu}_i^2+\boldsymbol{\sigma}_i^2}{v}\right).
\end{aligned}
\end{equation}

Notably, in our implementation, we ignore the $\ln v$ term, because $\ln v$ is a constant, which will not influence the performance. Therefore, the final loss function in our implementation is:
\begin{equation}\label{eq:final:loss}
    D_{\text{KL}}(\mathcal{N}_1\parallel\mathcal{N}_2) \simeq \mathcal{L}_{\text{KL}}(\boldsymbol{x}) = \frac{1}{2}\sum_{i=1}^{d}\left(-\ln\boldsymbol{\sigma}_i^2 -1 + \frac{\boldsymbol{\mu}_i^2+\boldsymbol{\sigma}_i^2}{v}\right).
\end{equation}

Then, we choose the hyperparameter of loss function and optimization strategy in Equation~\ref{eq:final:loss} for both phenomena: 
\begin{enumerate}
    \item \textbf{Maximization Strategy (PCA+):}

For objective 1, we decide to apply the diffusion collapse, which can be translate to a optimization problem:
\begin{equation}
\max_{\boldsymbol{\delta}} \mathcal{L}_{\mathrm{KL}}(\boldsymbol{x} + \boldsymbol{\delta}) \quad \text{s.t.} \quad \|\boldsymbol{\delta}\|_\infty < \epsilon.\label{eq:attack+}
\end{equation}
For the KL divergence, we choose \(v = 1\), which align with the training process of VAE. The loss is already shown in Equation \ref{pca+}.
Analysis shows that maximizing this leads to:
\begin{equation}
|\boldsymbol{\mu}_i| \rightarrow \infty \quad  \text{and} \quad \sigma_i^2 \rightarrow\infty\quad \text{for all } i,
\end{equation}
resulting in diffusion collapse.

PCA+ optimization pushes the latent distribution $\mathcal{N}(\boldsymbol{\mu},\boldsymbol{\sigma}^2)$ away from the prior distribution $\mathcal{N}(\mathbf{0},\mathbf{I})$, effectively creating an out-of-distribution (OOD) scenario.
Notably, this protection task benefits from detectability unlike evasion-focused adversarial attacks, i.e., detection cannot remove perturbations but may deter infringers, while editing disruption remains effective.
This divergence introduces a significant domain shift for the downstream diffusion model, moving the data distribution into regions where the model's knowledge is limited or inaccurate.
Consequently, the diffusion model struggles to perform proper inference on the transformed data, as its fundamental assumptions about the data distribution are violated.

This scenario can be conceptualized as an expansion of the latent space, where data is forced to extend beyond the normal domain of the downstream diffusion model's understanding.
This manipulation can be seen as a method to ``confuse'' the diffusion model by presenting it with data that falls outside its expected input distribution. The effect is analogous to creating a domain shift in the context of VAEs, where the latent space is manipulated to fall outside the expected distribution that the decoder was trained on.

Intuitively the downstream LDM will pay attention to move those OOD points into the known distribution, causing results consistent with Objective 1 expectations.

    \item \textbf{Minimization Strategy (PCA-):}

For objective 2, we decide to apply concentration collapse, which can be consider as a optimization problem:
\begin{equation}
\min_{\boldsymbol{\delta}} \mathcal{L}_{\text{KL}}(\boldsymbol{x} + \boldsymbol{\delta}) \quad \text{s.t.} \quad \|\boldsymbol{\delta}\|_\infty < \epsilon.\label{eq:attack-}
\end{equation}
For this case, we choose \(v \rightarrow 0^+\) for the KL divergence. The loss is shown in Equation ~\ref{pca-} and the dominant term becomes:
\begin{equation}
\frac{\boldsymbol{\mu}_i^2 + \boldsymbol{\sigma}_i^2}{v}.
\end{equation}
Minimizing this expression leads to:
\begin{equation}
\boldsymbol{\mu}_i \rightarrow 0 \quad \text{and} \quad \boldsymbol{\sigma}_i^2 \rightarrow 0 \quad \text{for all } i,
\end{equation}
resulting in concentration collapse.

PCA- forces the posterior distribution to converge towards $\mathcal{N}(\mathbf{0},\mathbf{0})$, effectively collapsing the distribution into a point mass at $\mathbf{z}=0$, a Dirac delta function centered at the origin of the latent space.  In this context,  as the distribution collapses, all latent representations are concentrated at $\mathbf{z}=0$, leading to a severe loss of dimensionality in the latent space. The normal distribution $\mathcal{N}(\mathbf{0},\mathbf{0})$ ceases to be a valid probability distribution for describing the data, instead becoming a degenerate case where the distribution has zero variance in all dimensions.

When this collapsed distribution is passed through a LDM, it results in a complete breakdown of semantic information in the edited images. The LDM, expecting a meaningful distribution in the latent space, fails to generate coherent outputs from this degenerate input.

This collapse represents a total loss of information about the original data variability, matching the Objective 2, which makes it impossible for the LDM to recover meaningful features or structures.
\end{enumerate}

\subsection{Optimization Framework}
To craft adversarial samples, we use projected sign gradient ascent~\cite{madry2017towards} for iterative updates.
Our approach is inspired by a counterintuitive finding from \cite{sds}: they achieved better results by minimizing their loss function, which contradicts the intuitive expectation that maximization would be more effective.
Following this insight, we explore both gradient directions by altering the sign of our final loss function, which will be discussed in appendix. Our update process follows the form:
\begin{equation}
\boldsymbol{x}^{t+1}=\mathcal{P}_{\ell_{\infty}\left(\boldsymbol{\delta}^t\right)}^{\epsilon}\left(\boldsymbol{x}^t \pm \alpha \text{sign}\nabla_{\boldsymbol{x}^t}\mathcal{J}\left(\boldsymbol{x}^t\right)\right),
\end{equation}
where we apply plus for maximization and minus for minimization, and $\mathcal{P}_{\ell_{\infty}\left(\boldsymbol{\delta}^t\right)}^{\epsilon}\left(\cdot\right)$ will apply the projection on $\boldsymbol{\delta}^t$ in the $\epsilon$-ball of $\ell_{\infty}$ norm, $\alpha$ is a hyperparameter to adjust the learning rate during the optimization.

\input{float/algo}

Algorithm~\ref{alg:pca} presents the complete optimization procedure of our Posterior Collapse Attack. The algorithm takes a clean image $\boldsymbol{x}$ and the VAE encoder $\mathcal{E}$ as inputs, along with standard adversarial attack parameters: perturbation budget $\epsilon$, step size $\alpha$, and number of iterations $T$. Two method-specific parameters control the attack behavior: the variance parameter $v$ and the objective type $\text{obj} \in \{+, -\}$.

The optimization follows an iterative gradient-based approach. At each iteration $t$, we:
\begin{enumerate}
    \item Extract the mean $\boldsymbol{\mu}^t$ and variance $\boldsymbol{\sigma}^t$ from the encoder's output (Line 3).
    \item Compute the loss $\mathcal{L}_{\text{KL}}$ (in Equation~\ref{eq:final:loss}) between the posterior distribution $q(\boldsymbol{z}|\boldsymbol{x}^{t-1})$ and target distribution $p^*(\boldsymbol{z}) = \mathcal{N}(\mathbf{0}, v\mathbf{I})$ (Line 5).
    \item Calculate the gradient with respect to the input image (Line 7).
    \item Update the perturbation based on the objective type: gradient ascent for PCA+ ($\text{obj}=+$) to induce diffusion collapse, or gradient descent for PCA- ($\text{obj}=-$) to induce concentration collapse (Lines 8-13).
    \item Project the perturbation into the $\ell_\infty$ ball to satisfy the imperceptibility constraint (Line 15).
    \item Update the adversarial image while ensuring pixel values remain valid (Line 17).
\end{enumerate}

For PCA+, we set $v=1$ to align with the standard normal prior used in VAE training, causing the posterior to diverge and create an out-of-distribution scenario. For PCA-, we set $v=1\times10^{-8}$ to approximate zero variance, forcing the posterior to collapse to a degenerate point mass. This unified framework demonstrates the flexibility of our approach in achieving both protection objectives through a single algorithmic structure with different hyperparameter configurations.

Importantly, since our loss function $\mathcal{L}_{\text{KL}}$ operates only on the VAE encoder output ($\boldsymbol{\mu}, \boldsymbol{\sigma}^2$), the proposed method requires no knowledge of text prompts, providing a practical advantage in real-world scenarios where the infringer's editing intent is unknown.

\subsection{Transferability Analysis}

A key advantage of our method lies in its better transferability across different LDM architectures. This transferability stems from three fundamental design principles:

\textbf{Minimal Model Dependency.} Unlike existing methods that require white-box access to the major component of LDM, U-Net, our approach only requires access to the VAE encoder, which constitutes less than 4\% of the LDM parameters. This drastically reduces the dependency on model-specific knowledge.

\textbf{Architectural Commonality.} The VAE encoder serves as a universal bottleneck across various LDM architectures. Most state-of-the-art LDMs, including different versions of Stable Diffusion (SD1.4, SD1.5, SD2.0, etc), share similar VAE architectures with only minor variations. By targeting this common component, our attack naturally transfers across these models.

\textbf{Distribution-level Attack.} Rather than attacking specific learned features or model parameters, our method manipulates the fundamental statistical properties ($\boldsymbol{\mu}$ and $\boldsymbol{\sigma}$) of the VAE's posterior distribution. This distribution-level attack is inherently more robust to architectural variations, as any VAE-based system must process these statistical representations.

Consequently, adversarial perturbations crafted using one model's VAE encoder (e.g., SD1.5) can effectively transfer to other LDM variants without requiring additional knowledge of the target model's U-Net. This near-black-box setting makes our method practically applicable in real-world scenarios where full model access is unavailable.

\input{float/x-fxadv}
\input{float/fx-fxadv}

\section{Experiments Setup}
\subsection{Dataset}
In our experiments, we utilized a 1000-image-subset of the ImageNet~\cite{deng2009imagenet}, as selected by \cite{lin2020nesterov}. This choice aligns with established conventions in adversarial attack research.
All images are $512 \times 512$, ensuring consistency across evaluations.

\subsection{Baselines}
We compared our approach against several state-of-the-art methods, AdvDM~\cite{advdm}, PhotoGuard (PG)~\cite{photoguard}~\footnote{We followed the choice of \cite{sds}, using the most powerful variant of PG (diffusion attack).}, MIST~\cite{mist}, and SDS~\cite{sds}.
Those methods can be summarized into two groups: AdvDM, SDS, and PCA+ is for Objective 1; PG, MIST, and PCA- is for Objective 2.
We fix fix $\alpha=2$, $T = 40$ and $\epsilon=16$ for all methods, their implementations are followed with the unified framework~\footnote{\url{https://github.com/xavihart/Diff-Protect}} provided by SDS~\cite{sds}, where the rest hyperparameters left default.
We implement our method with $v=1$ for PCA+, $v=1 \times 10^{-8}$ for PCA-, the optimization follows the goal defined in Equation~\ref{eq:attack+} and Equation~\ref{eq:attack-}.

\subsection{Victim Models}
Our experiments primarily focused on popular LDMs, specifically the lightweight Stable Diffusion 1.4 (SD14) and 1.5 (SD15).
These models were chosen due to their widespread accessibility and higher potential for misuse.

To assess the transferability across different resolution and architectures, we also tested the first 100 images on Stable Diffusion 2.0 (SD20) with $768\times768$, and Stable Diffusion XL (SDXL) with $1024\times1024$ for all methods.
All reported results of our method in transferability leverage the VAE of SD15 as the surrogate model.

\subsection{Varied Image Editing Prompts}
To further evaluate the robustness and versatility of our method, we conducted inference using a diverse set of prompts. These prompts include:
\begin{itemize}
    \item P1: an empty prompt.
    \item P2: a caption generated by the vision-language model~\cite{li2022blip}.
    \item P3: \emph{``add some snow''} to represent weather modifications.
    \item P4: \emph{``apply sunset lighting''} to test lighting adjustments.
    \item P5: \emph{``make it like a watercolor painting''} to examine style transfer capabilities.
\end{itemize}

To evaluate our method's robustness across diverse real-world editing scenarios, we additionally constructed a dataset of 200 image-prompt pairs from LAION~\cite{schuhmann2022laion} using Claude's multimodal API\footnote{claude-sonnet-4-5-20250929}. The dataset covers five editing categories: Facial Expression Change, Object Addition, Object Removal, Pose Transformation, and Background Replacement, with prompts automatically generated based on image content.

Through this varied set of prompts, we aimed to assess our method's effectiveness across a wide range of potential image manipulation scenarios.

\begin{figure*}[t]
    \centering
    \includegraphics[width=\linewidth]{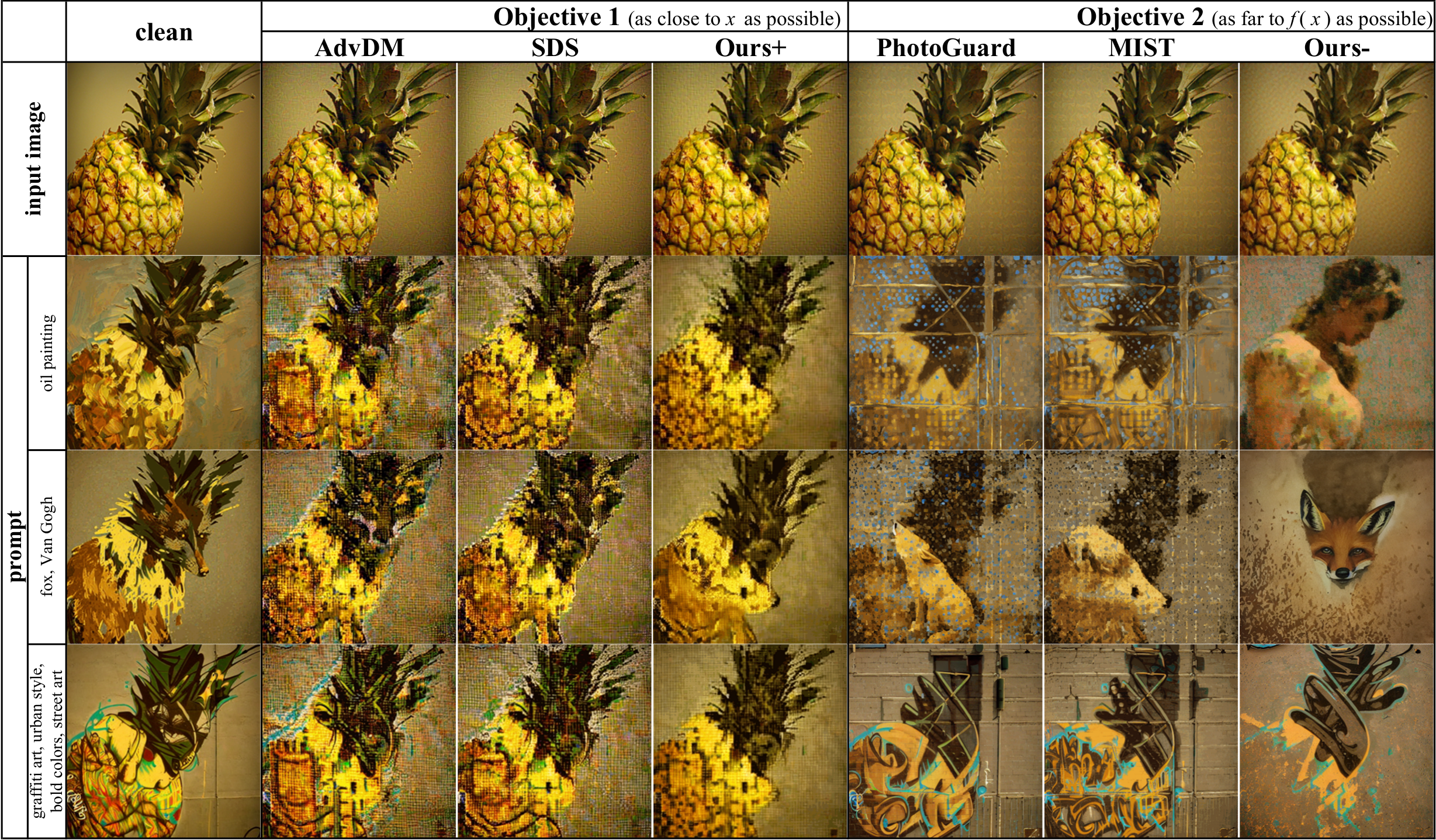}
    \caption{Qualitative comparison of image editing results. The first row is the original image and protected images. For the rest, each row shows different methods applied to a pineapple image under various editing prompts and protection objectives. Notably, $x$ refers to the original image will be edited by LDM; $f(\cdot)$ refers to the LDM-based image editing.}
    \label{fig:demo1}
\end{figure*}

\begin{figure*}[t]
    \centering
    \includegraphics[width=\linewidth]{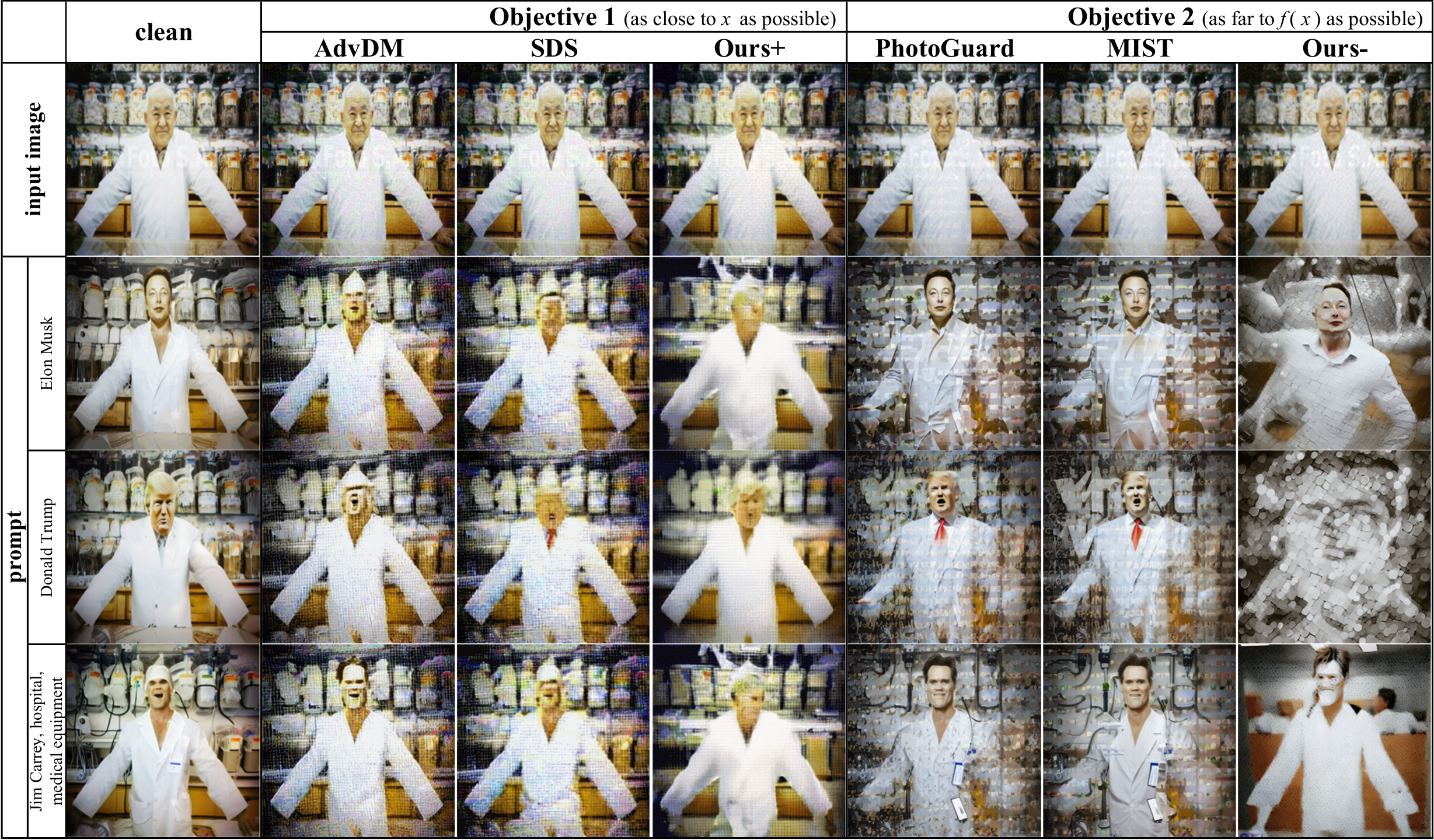}
    \caption{Qualitative comparison of image editing results. The first row is the original image and protected images. For the rest, each row shows different methods applied to a portrait photo under various editing prompts and protection objectives. Notably, $x$ refers to the original image will be edited by LDM; $f(\cdot)$ refers to the LDM-based image editing.}
    \label{fig:demo2}
\end{figure*}

\subsection{Image Quality Assessment (IQA)}
To quantitatively assess the quality of the edited images pre- and post-attack, we employed five different IQAs:
\begin{itemize}
    \item \textbf{Peak Signal-to-Noise Ratio (PSNR)}: provides insight into the overall distortion introduced.\footnote{Range set to 1, aligning with PyTorch's image transformation.}
    \item \textbf{Fr\'echet Inception Distance (FID)}: offers a measure of the perceptual quality and generation diversity~\cite{heusel2017gans}.\footnote{\url{https://pytorch.org/torcheval/stable/}.}
    \item \textbf{Structural Similarity Index (SSIM)}: captures changes in local patterns of pixel intensities~\cite{wang2004image}.\footnote{Gaussian kernel size is 11, \url{https://github.com/Po-Hsun-Su/pytorch-ssim}.}
    \item \textbf{Learned Perceptual Image Patch Similarity (LPIPS)}: measures human perceptual similarities~\cite{zhang2018unreasonable}.\footnote{Encoder is VGG, \url{https://github.com/richzhang/PerceptualSimilarity}.}
    \item \textbf{Aesthetic Color Distance Metric (ACDM)}: evaluates the overall color consistency~\cite{guo2024artwork,guo2026artwork}.\footnote{Gaussian kernel is 11, \url{https://github.com/ZhongliangGuo/ACDM}.}
\end{itemize}

\subsection{Evaluation Protocol}
Since baselines serve different objectives, for a comprehensive performance evaluation, we employ two distinct approaches.
for methods subject to Objective 1, we compute IQAs between clean images and edited adversarial samples, represented as $\operatorname{IQA}(\boldsymbol{x},f(\boldsymbol{x}_{adv}))$.
for methods subject to Objective 2, we calculate IQAs between edited clean images and edited adversarial samples, denoted as $\operatorname{IQA}(f(\boldsymbol{x}),f(\boldsymbol{x}_{adv}))$.

The interpretation of IQA metrics differs between these objectives: for Objective 1, IQA metrics approaching their optimal values indicate successful protection against unauthorized editing, while for Objective 2, IQA metrics deviating from their optimal values suggest better effectiveness in disrupting unauthorized editing attempts.

\subsection{Against Defending Methods}
To evaluate our attack's robustness against potential defenses, we first examined Adv-Clean~\cite{advcleaner}. This defense includes multi-stage transformation including image degradations and filter blurs, which designed to remove adversarial perturbations against diffusion models. The method inherently addresses common image modifications that occur during real-world distribution and manipulation.
Additionally, we evaluated these baseline methods against DiffPure, a pixel-level diffusion model-based adversarial purification method, to further assess their resilience against adversarial purification techniques.
We apply these defense on our attacked images and evaluate on adversarial samples surrogated by SD15.

\section{Results}

Throughout this section, ``PCA-'' and ``ours-'' refer to minimize the $\mathcal{L}_{\text{KL}}$ (consistent with Tab.~\ref{tab:x-fxadv}) with $v=1\times 10^{-8}$;
``PCA+'' and ``ours+'' refer to maximize the $\mathcal{L}_{\text{KL}}$ with $v = 1$.

In the subsequent tables, \textbf{bold text} denotes the best results,
arrows ($\uparrow$/$\downarrow$) indicate whether higher or lower values represent better protection performance.
``clean'' for each prompt is the IQA computed between clean image $\boldsymbol{x}$ and edited clean image $f(\boldsymbol{x})$.

\subsection{Image Editing Results with Varied Prompts}

\input{float/extra-prompts-obj1}

\input{float/extra-prompts-obj2}

Our extensive experimentation is consistent with the hypothesis: PCA+ is good at Objective 1, while PCA- excels in addressing Objective 2. The following sections delve deeper into these findings.

\subsubsection{Analysis of PCA+ Performance (Objective 1)}

Tab.~\ref{tab:x-fxadv} presents a comprehensive comparison of $\operatorname{IQA}(\boldsymbol{x},f(\boldsymbol{x}_{adv}))$ across various methods, highlighting the efficacy of our PCA+ approach.

The results demonstrate a consistent superior performance of our method across multiple Image Quality Assessment (IQA) metrics. Notably, PCA+ achieves the best scores in PSNR, FID, SSIM, and LPIPS across all evaluated scenarios, indicating a robust ability to maintain similarity between the edited adversarial image and the original image. This is further validated on the extended dataset (Tab.~\ref{tab:extended:obj1}) covering diverse editing categories including object removal and background replacement, where PCA+ maintains the best PSNR (19.06), FID (109.65), SSIM (0.3910), and LPIPS (0.6065).

This comprehensive performance across various metrics underscores the method's effectiveness in addressing Objective 1.

As evidenced in Fig.~\ref{fig:demo1} and Fig.~\ref{fig:demo2}, visual analysis reveals that PCA+ produces results visually comparable to established methods such as SDS and AdvDM. A characteristic feature observed in the edited images is the appearance of fine grid artifacts, which is consistent across these methods.

However, PCA+ distinguishes itself through enhanced resistance to undesired edits.

This superiority is not only reflected in the quantitative metrics but also apparent in qualitative analysis.

A striking example is observed in the last two rows of Fig.~\ref{fig:demo1}, where PCA+ successfully prevents the appearance of fox head in the dark area (penultimate row) and Graffiti-like teal artifacts (last row) in $f(\boldsymbol{x}_{adv})$, a feat not achieved by other methods.

In Fig.~\ref{fig:demo2}, compared to other methods, PCA+ prevent the generation of the target facial details that infringers want to make.

All visualizations showcase that the PCA+ can effectively prevent the unauthorized image editing by LDMs.

\subsubsection{Analysis of PCA- Performance (Objective 2)}

Tab.~\ref{tab:fx-fxadv} provides a detailed comparison of $\operatorname{IQA}(f(\boldsymbol{x}),f(\boldsymbol{x}_{adv}))$ across different methods, showcasing the effectiveness of our PCA- approach in addressing Objective 2. The results reveal that PCA- achieves the highest level of disruption to the infringer-desired output among all evaluated methods. On the extended dataset (Tab.~\ref{tab:extended:obj2}), PCA- demonstrates superior disruption with the lowest PSNR (13.92), SSIM (0.3276), and highest LPIPS (0.6468) and ACDM (0.1413) across challenging scenarios.

Visual analysis of the results, as illustrated in Fig.~\ref{fig:demo1} and Fig.~\ref{fig:demo2}, demonstrates a striking contrast between PCA- and other methods. Alternative approaches tend to preserve some degree of structural information similar to $f(\boldsymbol{x})$ in their $f(\boldsymbol{x}_{adv})$, however, PCA- stands out by completely obliterating the semantic information of the original image.

A concrete example of this capability is evident in Fig.~\ref{fig:demo1}. Under the prompt ``fox, Van Gogh,'' other methods still maintain recognizable structural elements of the pineapple, including the outlines of both leaves and fruit portions. In contrast, PCA- almost completely deconstructs this structural information, producing images that appear more randomly generated rather than edited from the original image. Furthermore, in the last row, while other methods preserve the wall seams characteristic of normal outputs, our method not only eliminates these seams but also successfully removes the contours of the pineapple fruit section in the edited results.

Fig.~\ref{fig:demo2} further demonstrates the effectiveness of PCA- across different image categories. Under various prompts, our method consistently achieves greater disruption to the generated images, pushing them further away from infringer-desired images. This is particularly evident when the prompt is ``Donald Trump'', where the semantic content of the original image is barely preserved, resulting in unrecognizable outputs: a level of disruption that other methods fail to achieve.

\subsection{Transferability}

\begin{figure*}[t]
\revisedblock
    \centering
    \includegraphics[width=\linewidth]{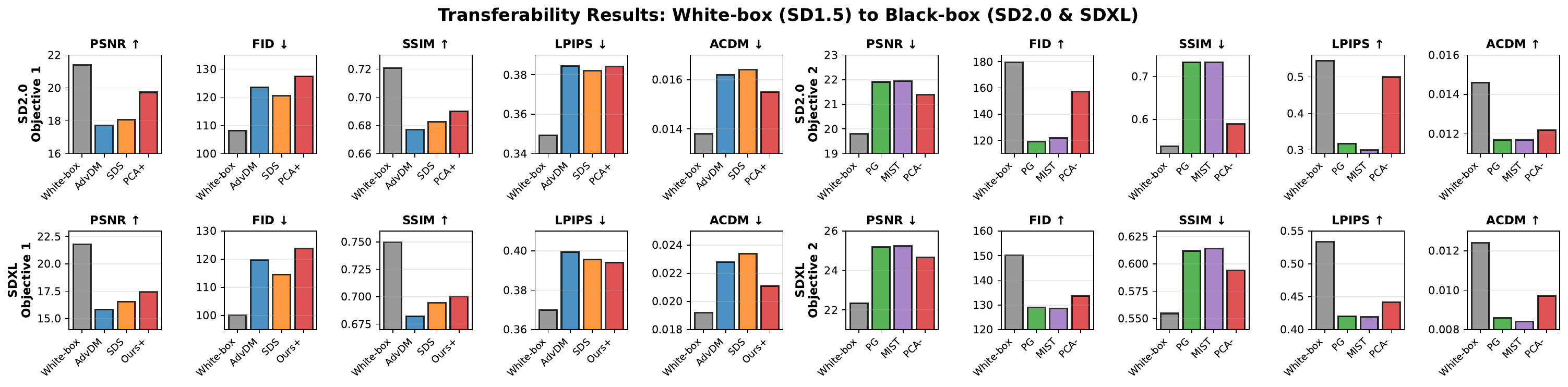}
    \caption{Transferability results on SD2.0 and SDXL. White-box uses the target model's VAE; black-box uses SD1.5's pretrained weights as surrogate. (Left) Objective 1 results. (Right) Objective 2 results. $\uparrow/\downarrow$ indicate higher/lower is better. All subplots share the same axes: x-axis shows baseline methods; y-axis shows IQA scores.}
    \label{fig:transfer}
\end{figure*}

As illustrated in Fig.~\ref{fig:transfer}, our method demonstrates better transferability in achieving both protection objectives. For methods subject to Objective 1, our approach consistently achieves optimal results in PSNR, SSIM, and ACDM metrics when transferred to either SD20 or SDXL, achieving comparable performance in a black-box manner without accessing any target model parameters. Similarly, for Objective 2, our method maintains superior performance across both SD2.0 and SDXL architectures.

Notably, these performance gains are achieved with reduced model parameter access, lower runtime requirements, and decreased VRAM consumption. This suggests that our method can effectively prevent unauthorized image editing across Stable Diffusion-based models by leveraging an inherent architectural vulnerability, Posterior Collapse, in a more generalizable manner.

\begin{figure}[t]
    \centering
    \includegraphics[width=0.49\linewidth]{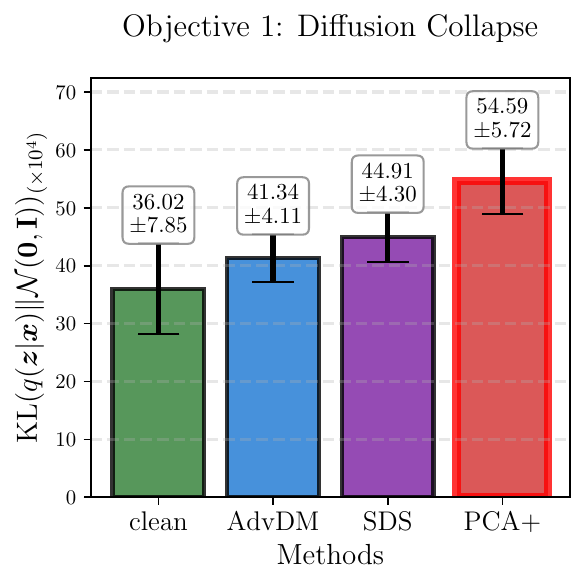}
    \includegraphics[width=0.492\linewidth]{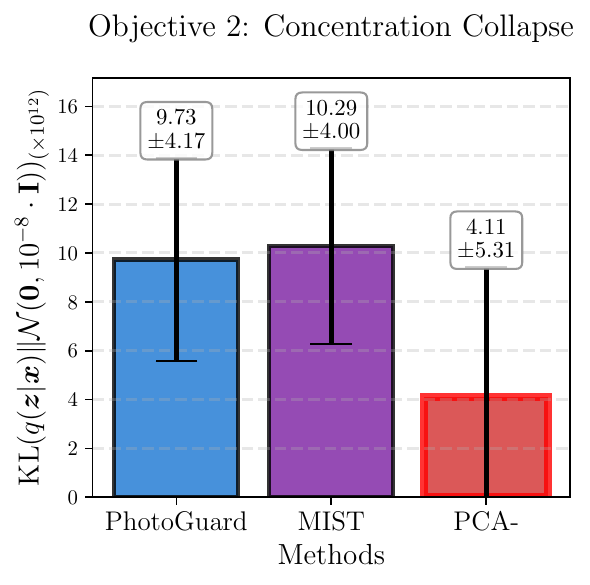}
    \caption{Analysis of KL divergence. (left) Methods subject to Objective 1, \textbf{Higher values indicate stronger deviation} from the standard prior. (right) Methods subject to Objective 2, \textbf{Lower values indicate stronger concentration} toward the origin, approaching a Dirac $\delta$ function. Values are reported as mean with std as error bar.}
    \label{fig:kl:obj12}
\end{figure}
\subsection{Analysis of Collapse Phenomena}

To validate our claims about the two types of collapse in the Section~\ref{sec:detailed-collapses}, we measured KL divergence statistics for each method grouped by Objectives 1 and 2, with results shown in Fig.~\ref{fig:kl:obj12}. PCA+ achieves the highest KL divergence, approximately 32\% above the second-best method, empirically confirming it creates the strongest out-of-distribution effect by pushing $q(\boldsymbol{z}|\boldsymbol{x})$ furthest from $\mathcal{N}(\boldsymbol{0},\mathbf{I})$. Conversely, PCA- demonstrates the lowest KL, representing a 58\% reduction compared to baselines, confirming it most effectively collapses the distribution toward a Dirac $\delta$ function.

\subsection{Against Defense}
As shown in Tab.~\ref{tab:defense:advclean}, we evaluate the robustness against Adv-Clean, which employs comprehensive low-level vision processing operations. For Objective 1, our method demonstrates superior resistance to various image degradations and filtering operations, maintaining better image preservation metrics (PSNR and SSIM) compared to baselines. For Objective 2, our approach shows stronger capability in preventing unauthorized edits, achieving significantly larger deviations from expected editing results across all metrics, particularly in FID and ACDM. This indicates our methods effectively preserve the capability to influence the image editing even under multiple image transformations including degradations and filtering blur.

The evaluation against DiffPure is shown in Tab.~\ref{tab:defense:diffpure}, which implements a diffusion model-based adversarial purification mechanism. For Objective 1, our approach outperforms baseline methods in preserving original image content, achieving higher scores in key image quality metrics. For Objective 2, our method maintains stronger protective capabilities compared to PhotoGuard and MIST, showing larger deviations from target editing effects. These results demonstrate that our method better preserves its protective strength against learning-based purification attempts compared to existing approaches.

This comprehensive evaluation suggests that our method maintains its effectiveness under both protection objectives even in the presence of sophisticated defense mechanisms, whether they employ traditional image processing operations or learning-based purification approaches. Notably, these superior protective capabilities are achieved with reduced computational requirements: lower VRAM consumption, faster runtime, and minimal target model parameter access, demonstrating effective protection in an almost black-box manner.

\input{float/defense-adv-clean}
\input{float/defense-diffpure}

\section{Hyperparameters Analysis}
We conduct comprehensive analysis on all hyperparameters. It is worth noting that, for the hyperparameter $v$, we do the analysis with evaluation protocols subject to both Objective 1 and Objective 2, which aims to evidence our choice of $v=1$ for PCA+ and $v=10^{-8}$ for PCA-. Once the $v$ is determined, the rest analysis will follow the evaluation protocol of Objective 2 to find the optimal value.

\subsection{Analysis on $v$}

We conduct a analysis on the variance hyperparameter $v$ in our attack target distribution $p^{\ast}(\boldsymbol{z})=\mathcal{N}(\mathbf{0},v\mathbf{I})$. This hyperparameter controls the extent of the latent space manipulation in our attack.

We compare two values, $v=1$ and $v=1\times 10^{-8}$, for both PCA+ and PCA- methods. The choice of $v=10^{-8}$ is to approximate a variance close to 0 while maintaining numerical stability.

\subsubsection{PCA+}
\input{float/ablation-v-objs}
For PCA+, our objective is to maintain consistency with the original image while still preventing unauthorized edits. As shown in Tab.~\ref{tab:abl:v:obj1}, the performance of PCA+ is remarkably similar for both $v=1$ and $v=1\times 10^{-8}$ across most metrics.
However, the FID scores for $v=1$ are consistently slightly lower across all prompts and models.
The marginally superior FID performance of $v=1$ can be attributed to its alignment with the standard normal distribution, which is commonly used as the prior in many diffusion models.
Based on this observation, we choose $v=1$ for our main experiments with PCA+.

\subsubsection{PCA-}
As shown in Tab.~\ref{tab:ablation:v}, setting $v$ close to 0 outperforms $v = 1$ across all prompts and both models, as evidenced by all five evaluated metrics.
This consistent improvement can be attributed to the tighter concentration of the target distribution around zero when $v$ is smaller. A smaller $v$ encourages the posterior distribution to collapse more severely, leading to more significant disruption in the latent space and, consequently, in the generated images.
Our choice of $v = 10^{-8}$ in the main experiments reflects this finding.
\input{float/ablation-v}
\input{float/ablation-alpha}
\subsection{Analysis on $\alpha$}

We conducted a analysis on the learning rate $\alpha$ used in our projected sign gradient ascent. We experimented with $\alpha$ values of 1, 2, and 4 across five different prompts (P1 to P5) and two Stable Diffusion models (SD14 and SD15). Tab.~\ref{tab:ablation:alpha} presents the results of this study.
As shown in Tab.~\ref{tab:ablation:alpha}, $\alpha = 2$ generally yields the best performance across most prompts and metrics. This trend is consistent across different prompts and both models, with a few exceptions where $\alpha = 1$ or $\alpha = 4$ perform slightly better in specific metrics.
The superior performance of $\alpha = 2$ suggests that this learning rate provides an optimal balance between the step size and the number of iterations in our attack algorithm. It allows for sufficiently large updates to effectively perturb the latent space while avoiding overshooting that might occur with larger learning rates.

\subsection{Analysis on $T$}
\begin{figure*}[t]
    \revisedblock
    \centering
    \includegraphics[width=\linewidth]{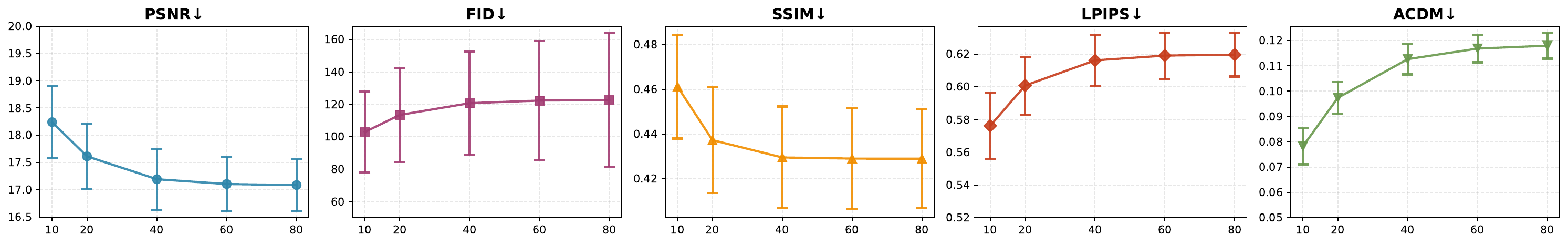}
    \caption{Hyperparameters Analysis on $T$. Performance metrics averaged over five prompts (P1-P5). Error bars indicate standard deviation. All subplots share the same axes: x-axis shows attack steps $T$; y-axis shows IQA scores.}
    \label{fig:ablation:t}
\end{figure*}

To investigate the impact of the number of attack steps on our method's performance, we conducted a analysis varying the attack steps $T$. We experimented with $T$ values of 10, 20, 40, 60, and 80 across five different prompts (P1 to P5) on SD15. Fig.~\ref{fig:ablation:t} presents the results of this study.

As shown in Fig.~\ref{fig:ablation:t}, the performance of our method improves substantially in early iterations and reaches practical convergence around $T=40$. Specifically, from $T=10$ to $T=40$, we observe significant improvements across all metrics (e.g., PSNR decreases by 1.13, FID increases by approximately 15-20). However, beyond $T=40$, the marginal gains diminish considerably, increasing from $T=40$ to $T=80$ yields less than 5\% relative improvement in all metrics, with overlapping error bars indicating statistical insignificance. This convergence behavior demonstrates that $T=40$ provides an effective balance between attack performance and computational efficiency.

\subsection{Hyperparameters analysis on $\epsilon$}
\begin{figure*}[t]
\revisedblock
    \includegraphics[width=\linewidth]{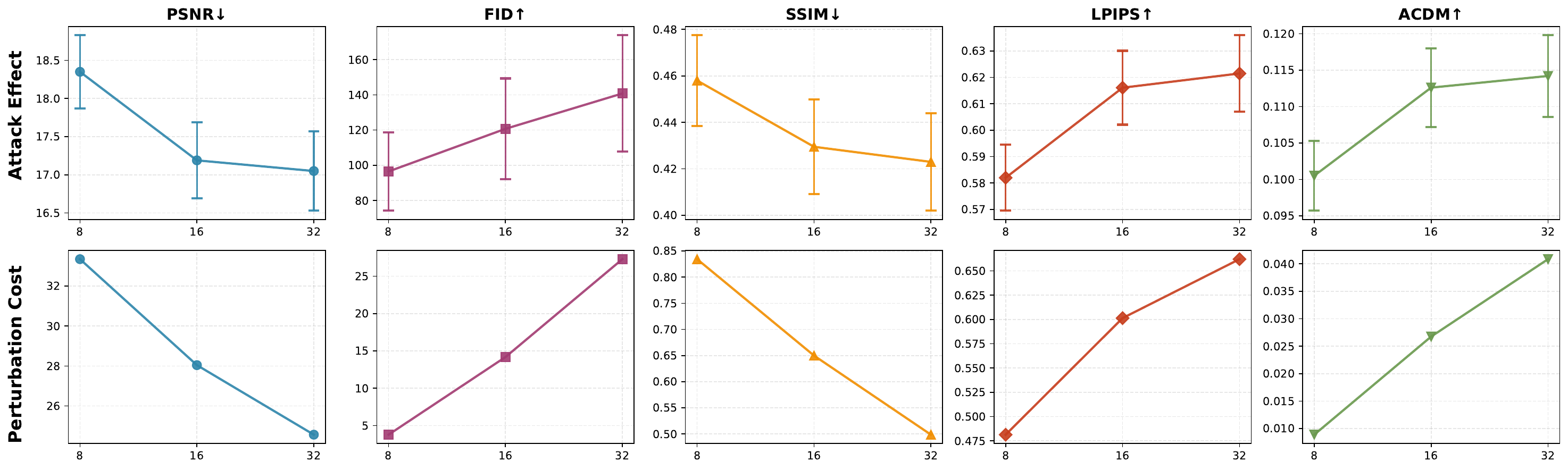}
    \caption{Hyperparameters analysis on $\epsilon$. (Top) Attack effectiveness reported in mean\textpm std across five prompts. (Bottom) Perturbation cost, measured as the visual quality degradation between original images $\boldsymbol{x}$ and adversarial samples $\boldsymbol{x}_{adv}$ across five IQA metrics. Each subplot shows how the corresponding metric changes with different $\epsilon$ values (x-axis: $\ell_\infty$ budget; y-axis: IQA score).}
    \label{fig:ablation:eps}
\end{figure*}

We evaluate our method across different perturbation budgets ($\epsilon = 8, 16, 32$) to analyze the trade-off between attack effectiveness and perturbation visibility. As shown in Fig.~\ref{fig:ablation:eps}, increasing $\epsilon$ from 8 to 16 yields substantial improvement in attack effectiveness with moderate perturbation cost. However, further increasing to $\epsilon = 32$ provides minimal additional gain while significantly degrading image quality. This diminishing return pattern validates $\epsilon = 16$ as the balanced operating point, which aligns with the standard perturbation budget widely adopted in prior adversarial attack literature~\cite{advdm,photoguard,mist,sds}.

\section{Conclusion}
In this paper, we propose the Posterior Collapse Attack (PCA), a novel protection method against unauthorized LDM-based image editing. PCA requires less than 4\% of the model parameters, enabling faster operation and lower VRAM consumption compared to existing approaches. Furthermore, our method demonstrates better transferability across different VAE-based Stable Diffusion architectures.

By extending the study of Posterior Collapse, an architecture-level vulnerability in VAE training, to the inference phase, we identify two distinct phenomena: diffusion collapse and concentration collapse, each serving different protection objectives. We introduce a novel loss function targeting the VAE encoder of LDMs, which can induce different types of collapses in the VAE's posterior distribution through single hyperparameter adjustment.

Experimental results demonstrate that both variants of PCA significantly outperform previous methods in different protection objective. Our work contributes to the ongoing efforts to secure digital assets in an era of rapidly advancing generative AI, while addressing the socio-technical challenges posed by malicious AI misuse.

\clearpage
\bibliography{reference}
\bibliographystyle{IEEEtran}
\clearpage

\end{document}

%% file: float/algo.tex
\begin{algorithm}[t]
\caption{Posterior Collapse Attack (PCA)}
\label{alg:pca}
\begin{algorithmic}[1]
\REQUIRE Clean image $\boldsymbol{x}$, VAE encoder $\mathcal{E}$, perturbation budget $\epsilon$, step size $\alpha$, number of iterations $T$, variance parameter $v$, objective type $\text{obj} \in \{+, -\}$
\ENSURE Adversarial image $\boldsymbol{x}_{\text{adv}}$

\STATE Initialize: $\boldsymbol{x}^{0} \leftarrow \boldsymbol{x}$, $\boldsymbol{\delta}^{0} \leftarrow \mathbf{0}$

\FOR{$t = 1$ to $T$}
    \STATE // Forward pass through VAE encoder
    \STATE $\boldsymbol{\mu}^t, \boldsymbol{\sigma}^t \leftarrow \mathcal{E}(\boldsymbol{x}^{t-1})$
    
    \STATE // Compute the loss (in Equation~\ref{eq:final:loss})
    \STATE $\mathcal{L}_{\text{KL}}(\boldsymbol{x}^{t-1}) = \frac{1}{2}\sum_{i=1}^{d}\left(-\ln(\boldsymbol{\sigma}_i^t)^2 - 1 + \frac{(\boldsymbol{\mu}_i^t)^2 + (\boldsymbol{\sigma}_i^t)^2}{v}\right)$
    
    \STATE // Compute gradient
    \STATE $\boldsymbol{g}^t \leftarrow \nabla_{\boldsymbol{x}^{t-1}} \mathcal{L}_{\text{KL}}(\boldsymbol{x}^{t-1})$
    
    \IF{$\text{obj} = +$}
        \STATE // Maximize for diffusion collapse (PCA+)
        \STATE $\boldsymbol{\delta}^t \leftarrow \boldsymbol{\delta}^{t-1} + \alpha \cdot \text{sign}(\boldsymbol{g}^t)$
    \ELSIF{$\text{obj} = -$}
        \STATE // Minimize for concentration collapse (PCA-)
        \STATE $\boldsymbol{\delta}^t \leftarrow \boldsymbol{\delta}^{t-1} - \alpha \cdot \text{sign}(\boldsymbol{g}^t)$
    \ENDIF
    
    \STATE // Project perturbation into $\ell_\infty$ ball
    \STATE $\boldsymbol{\delta}^t \leftarrow \text{clip}(\boldsymbol{\delta}^t, -\epsilon, \epsilon)$
    
    \STATE // Update adversarial image
    \STATE $\boldsymbol{x}^t \leftarrow \text{clip}(\boldsymbol{x} + \boldsymbol{\delta}^t, 0, 1)$
\ENDFOR

\STATE $\boldsymbol{x}_{\text{adv}} \leftarrow \boldsymbol{x}^T$

\RETURN $\boldsymbol{x}_{\text{adv}}$
\end{algorithmic}
\end{algorithm}

%% file: float/x-fxadv.tex
\begin{table*}[ht]
\caption{Results on IQAs for Methods Subject to Objective 1.}
\label{tab:x-fxadv}
\centering
\begin{tabular}{cccccccccccc}
\hline
                                          &                                                    & \multicolumn{2}{c}{PSNR$\uparrow$}                                                                      & \multicolumn{2}{c}{FID$\downarrow$}                                                                       & \multicolumn{2}{c}{SSIM$\uparrow$}                                                                        & \multicolumn{2}{c}{LPIPS$\downarrow$}                                                                     & \multicolumn{2}{c}{ACDM$\downarrow$}                                                 \\
\multirow{-2}{*}{prompt}                  & \multirow{-2}{*}{method}                           & \multicolumn{1}{c}{SD14}                           & \multicolumn{1}{c}{SD15}                           & \multicolumn{1}{c}{SD14}                            & \multicolumn{1}{c}{SD15}                            & \multicolumn{1}{c}{SD14}                            & \multicolumn{1}{c}{SD15}                            & \multicolumn{1}{c}{SD14}                            & \multicolumn{1}{c}{SD15}                            & \multicolumn{1}{c}{SD14}                            & \multicolumn{1}{c}{SD15}       \\ \hline
\multicolumn{1}{c|}{}                     & \multicolumn{1}{c|}{clean}                         & 19.09                                              & 19.09                                              & 54.95                                               & 54.25                                               & 0.5911                                              & 0.5914                                              & 0.3948                                              & 0.3943                                              & 0.0243                                              & 0.0242                         \\ \cline{2-12} 
\multicolumn{1}{c|}{}                     & \multicolumn{1}{c|}{AdvDM}                         & \multicolumn{1}{c|}{16.93}                         & \multicolumn{1}{c|}{16.93}                         & \multicolumn{1}{c|}{{172.10}}                   & \multicolumn{1}{c|}{169.12}                         & \multicolumn{1}{c|}{0.2748}                         & \multicolumn{1}{c|}{0.2746}                         & \multicolumn{1}{c|}{0.5783}                         & \multicolumn{1}{c|}{0.5778}                         & \multicolumn{1}{c|}{\textbf{0.0503}}                & \textbf{0.0508}                \\
\multicolumn{1}{c|}{}                     & \multicolumn{1}{c|}{SDS}                           & \multicolumn{1}{c|}{{17.15}}                   & \multicolumn{1}{c|}{{17.12}}                   & \multicolumn{1}{c|}{172.47}                         & \multicolumn{1}{c|}{169.68}                         & \multicolumn{1}{c|}{0.2944}                         & \multicolumn{1}{c|}{0.2924}                         & \multicolumn{1}{c|}{{0.5687}}                   & \multicolumn{1}{c|}{{0.5690}}                   & \multicolumn{1}{c|}{{0.0523}}                   & {0.0530}                   \\
\multicolumn{1}{c|}{\multirow{-4}{*}{P1}} & \multicolumn{1}{c|}{Ours+}                         & \multicolumn{1}{c|}{\textbf{18.85}}                & \multicolumn{1}{c|}{\textbf{18.82}}                & \multicolumn{1}{c|}{\textbf{109.01}}                & \multicolumn{1}{c|}{\textbf{108.57}}                & \multicolumn{1}{c|}{\textbf{0.4424}}                & \multicolumn{1}{c|}{\textbf{0.4396}}                & \multicolumn{1}{c|}{\textbf{0.5349}}                & \multicolumn{1}{c|}{\textbf{0.5345}}                & \multicolumn{1}{c|}{0.0593}                         & 0.0601                         \\ \hline
\multicolumn{1}{c|}{}                     & \multicolumn{1}{c|}{clean}                         & 18.68                                              & 18.69                                              & 45.34                                               & 45.41                                               & 0.5829                                              & 0.5838                                              & 0.3993                                              & 0.3986                                              & 0.0253                                              & 0.0251                         \\ \cline{2-12} 
\multicolumn{1}{c|}{}                     & \multicolumn{1}{c|}{AdvDM}                         & \multicolumn{1}{c|}{16.67}                         & \multicolumn{1}{c|}{16.65}                         & \multicolumn{1}{c|}{103.37}                         & \multicolumn{1}{c|}{101.18}                         & \multicolumn{1}{c|}{0.2826}                         & \multicolumn{1}{c|}{0.2822}                         & \multicolumn{1}{c|}{0.5658}                         & \multicolumn{1}{c|}{0.5649}                         & \multicolumn{1}{c|}{\textbf{0.0530}}                & \textbf{0.0526}                \\
\multicolumn{1}{c|}{}                     & \multicolumn{1}{c|}{SDS}                           & \multicolumn{1}{c|}{{16.89}}                   & \multicolumn{1}{c|}{{16.85}}                   & \multicolumn{1}{c|}{107.53}                         & \multicolumn{1}{c|}{106.82}                         & \multicolumn{1}{c|}{0.3009}                         & \multicolumn{1}{c|}{0.2993}                         & \multicolumn{1}{c|}{{0.5587}}                   & \multicolumn{1}{c|}{{0.5585}}                   & \multicolumn{1}{c|}{{0.0555}}                   & {0.0555}                   \\
\multicolumn{1}{c|}{\multirow{-4}{*}{P2}} & \multicolumn{1}{c|}{Ours+}                         & \multicolumn{1}{c|}{\textbf{18.61}}                & \multicolumn{1}{c|}{\textbf{18.60}}                & \multicolumn{1}{c|}{\textbf{70.40}}                 & \multicolumn{1}{c|}{\textbf{70.08}}                 & \multicolumn{1}{c|}{\textbf{0.4460}}                & \multicolumn{1}{c|}{\textbf{0.4444}}                & \multicolumn{1}{c|}{\textbf{0.5276}}                & \multicolumn{1}{c|}{\textbf{0.5269}}                & \multicolumn{1}{c|}{0.0602}                         & 0.0608                         \\ \hline
\multicolumn{1}{c|}{}                     & \multicolumn{1}{c|}{clean}                         & 18.50                                              & 18.54                                              & 78.89                                               & 78.98                                               & 0.5722                                              & 0.5735                                              & 0.4284                                              & 0.4281                                              & 0.0268                                              & 0.0266                         \\ \cline{2-12} 
\multicolumn{1}{c|}{}                     & \multicolumn{1}{c|}{AdvDM}                         & \multicolumn{1}{c|}{16.54}                         & \multicolumn{1}{c|}{16.53}                         & \multicolumn{1}{c|}{170.08}                         & \multicolumn{1}{c|}{169.18}                         & \multicolumn{1}{c|}{0.2734}                         & \multicolumn{1}{c|}{0.2726}                         & \multicolumn{1}{c|}{0.5852}                         & \multicolumn{1}{c|}{0.5823}                         & \multicolumn{1}{c|}{\textbf{0.0538}}                & \textbf{0.0536}                \\
\multicolumn{1}{c|}{}                     & \multicolumn{1}{c|}{SDS}                           & \multicolumn{1}{c|}{{16.83}}                   & \multicolumn{1}{c|}{{16.79}}                   & \multicolumn{1}{c|}{{167.93}}                   & \multicolumn{1}{c|}{{167.95}}                   & \multicolumn{1}{c|}{0.2930}                         & \multicolumn{1}{c|}{0.2898}                         & \multicolumn{1}{c|}{{0.5755}}                   & \multicolumn{1}{c|}{{0.5737}}                   & \multicolumn{1}{c|}{{0.0560}}                   & {0.0560}                   \\
\multicolumn{1}{c|}{\multirow{-4}{*}{P3}} & \multicolumn{1}{c|}{Ours+}                         & \multicolumn{1}{c|}{\textbf{18.55}}                & \multicolumn{1}{c|}{\textbf{18.58}}                & \multicolumn{1}{c|}{\textbf{114.34}}                & \multicolumn{1}{c|}{\textbf{112.30}}                & \multicolumn{1}{c|}{\textbf{0.4348}}                & \multicolumn{1}{c|}{\textbf{0.4355}}                & \multicolumn{1}{c|}{\textbf{0.5420}}                & \multicolumn{1}{c|}{\textbf{0.5405}}                & \multicolumn{1}{c|}{0.0605}                         & 0.0605                         \\ \hline
\multicolumn{1}{c|}{}                     & \multicolumn{1}{c|}{clean}                         & 18.82                                              & 18.84                                              & 69.26                                               & 69.17                                               & 0.5886                                              & 0.5905                                              & 0.4116                                              & 0.4096                                              & 0.0256                                              & 0.0256                         \\ \cline{2-12} 
\multicolumn{1}{c|}{}                     & \multicolumn{1}{c|}{AdvDM}                         & \multicolumn{1}{c|}{16.65}                         & \multicolumn{1}{c|}{16.62}                         & \multicolumn{1}{c|}{165.67}                         & \multicolumn{1}{c|}{{162.57}}                   & \multicolumn{1}{c|}{0.2730}                         & \multicolumn{1}{c|}{0.2712}                         & \multicolumn{1}{c|}{0.5911}                         & \multicolumn{1}{c|}{0.5898}                         & \multicolumn{1}{c|}{\textbf{0.0567}}                & \textbf{0.0564}                \\
\multicolumn{1}{c|}{}                     & \multicolumn{1}{c|}{SDS}                           & \multicolumn{1}{c|}{{16.87}}                   & \multicolumn{1}{c|}{{16.85}}                   & \multicolumn{1}{c|}{{165.40}}                   & \multicolumn{1}{c|}{164.08}                         & \multicolumn{1}{c|}{0.2944}                         & \multicolumn{1}{c|}{0.2913}                         & \multicolumn{1}{c|}{{0.5798}}                   & \multicolumn{1}{c|}{{0.5786}}                   & \multicolumn{1}{c|}{{0.0590}}                   & {0.0587}                   \\
\multicolumn{1}{c|}{\multirow{-4}{*}{P4}} & \multicolumn{1}{c|}{Ours+}                         & \multicolumn{1}{c|}{\textbf{18.72}}                & \multicolumn{1}{c|}{\textbf{18.77}}                & \multicolumn{1}{c|}{\textbf{110.03}}                & \multicolumn{1}{c|}{\textbf{109.44}}                & \multicolumn{1}{c|}{\textbf{0.4447}}                & \multicolumn{1}{c|}{\textbf{0.4447}}                & \multicolumn{1}{c|}{\textbf{0.5396}}                & \multicolumn{1}{c|}{\textbf{0.5385}}                & \multicolumn{1}{c|}{0.0614}                         & 0.0609                         \\ \hline
\multicolumn{1}{c|}{}                     & \multicolumn{1}{c|}{clean}                         & 18.52                                              & 18.55                                              & 83.84                                               & 82.33                                               & 0.5724                                              & 0.5737                                              & 0.4321                                              & 0.4298                                              & 0.0268                                              & 0.0267                         \\ \cline{2-12} 
\multicolumn{1}{c|}{}                     & \multicolumn{1}{c|}{AdvDM}                         & \multicolumn{1}{c|}{16.54}                         & \multicolumn{1}{c|}{16.47}                         & \multicolumn{1}{c|}{180.73}                         & \multicolumn{1}{c|}{181.80}                         & \multicolumn{1}{c|}{0.2609}                         & \multicolumn{1}{c|}{0.2592}                         & \multicolumn{1}{c|}{0.5925}                         & \multicolumn{1}{c|}{0.5923}                         & \multicolumn{1}{c|}{\textbf{0.0559}}                & \textbf{0.0553}                \\
\multicolumn{1}{c|}{}                     & \multicolumn{1}{c|}{SDS}                           & \multicolumn{1}{c|}{{16.73}}                   & \multicolumn{1}{c|}{{16.61}}                   & \multicolumn{1}{c|}{182.40}                         & \multicolumn{1}{c|}{183.48}                         & \multicolumn{1}{c|}{0.2769}                         & \multicolumn{1}{c|}{0.2734}                         & \multicolumn{1}{c|}{{0.5855}}                   & \multicolumn{1}{c|}{{0.5860}}                   & \multicolumn{1}{c|}{{0.0588}}                   & {0.0585}                   \\
\multicolumn{1}{c|}{\multirow{-4}{*}{P5}} & \multicolumn{1}{c|}{Ours+}                         & \multicolumn{1}{c|}{\textbf{18.50}}                & \multicolumn{1}{c|}{\textbf{18.40}}                & \multicolumn{1}{c|}{\textbf{124.06}}                & \multicolumn{1}{c|}{\textbf{127.72}}                & \multicolumn{1}{c|}{\textbf{0.4144}}                & \multicolumn{1}{c|}{\textbf{0.4092}}                & \multicolumn{1}{c|}{\textbf{0.5441}}                & \multicolumn{1}{c|}{\textbf{0.5444}}                & \multicolumn{1}{c|}{0.0618}                         & 0.0640                         \\ \hline
\end{tabular}
\end{table*}

%% file: float/fx-fxadv.tex
\begin{table*}[ht]
\caption{Results on IQAs for Methods Subject to Objective 2.}
\label{tab:fx-fxadv}
\centering
\begin{tabular}{cccccccccccc}
\hline
                                          &                                                    & \multicolumn{2}{c}{PSNR$\downarrow$}                                                                      & \multicolumn{2}{c}{FID$\uparrow$}                                                                       & \multicolumn{2}{c}{SSIM$\downarrow$}                                                                        & \multicolumn{2}{c}{LPIPS$\uparrow$}                                                                     & \multicolumn{2}{c}{ACDM$\uparrow$}                                                 \\
\multirow{-2}{*}{prompt}                  & \multirow{-2}{*}{method}                           & \multicolumn{1}{c}{SD14}                           & \multicolumn{1}{c}{SD15}                           & \multicolumn{1}{c}{SD14}                            & \multicolumn{1}{c}{SD15}                            & \multicolumn{1}{c}{SD14}                            & \multicolumn{1}{c}{SD15}                            & \multicolumn{1}{c}{SD14}                            & \multicolumn{1}{c}{SD15}                            & \multicolumn{1}{c}{SD14}                            & \multicolumn{1}{c}{SD15}       \\ \hline
\multicolumn{1}{c|}{\multirow{3}{*}{P1}}        & \multicolumn{1}{c|}{PG}                            & \multicolumn{1}{c|}{{18.49}}                   & \multicolumn{1}{c|}{{18.52}}                   & \multicolumn{1}{c|}{\textbf{151.75}}                         & \multicolumn{1}{c|}{\textbf{147.85}}                         & \multicolumn{1}{c|}{0.4800}                         & \multicolumn{1}{c|}{0.4840}                         & \multicolumn{1}{c|}{{0.5647}}                   & \multicolumn{1}{c|}{{0.5629}}                   & \multicolumn{1}{c|}{{0.0786}}                   & {0.0792}                   \\
\multicolumn{1}{c|}{}                     & \multicolumn{1}{c|}{MIST}                          & \multicolumn{1}{c|}{18.50}                         & \multicolumn{1}{c|}{18.54}                         & \multicolumn{1}{c|}{150.20}                         & \multicolumn{1}{c|}{147.78}                         & \multicolumn{1}{c|}{0.4759}                         & \multicolumn{1}{c|}{0.4797}                         & \multicolumn{1}{c|}{0.5604}                         & \multicolumn{1}{c|}{0.5585}                         & \multicolumn{1}{c|}{0.0751}                         & 0.0757                         \\
\multicolumn{1}{c|}{}                     & \multicolumn{1}{c|}{Ours-}                         & \multicolumn{1}{c|}{\textbf{18.02}}                & \multicolumn{1}{c|}{\textbf{18.02}}                & \multicolumn{1}{c|}{139.11}                         & \multicolumn{1}{c|}{135.32}                         & \multicolumn{1}{c|}{\textbf{0.4441}}                   & \multicolumn{1}{c|}{\textbf{0.4453}}                   & \multicolumn{1}{c|}{\textbf{0.6188}}                & \multicolumn{1}{c|}{\textbf{0.6171}}                & \multicolumn{1}{c|}{\textbf{0.1106}}                & \textbf{0.1110}                \\ \hline
\multicolumn{1}{c|}{\multirow{3}{*}{P2}}  & \multicolumn{1}{c|}{PG}                            & \multicolumn{1}{c|}{{17.32}}                   & \multicolumn{1}{c|}{{17.31}}                   & \multicolumn{1}{c|}{\textbf{88.43}}                          & \multicolumn{1}{c|}{\textbf{91.00}}                          & \multicolumn{1}{c|}{0.4650}                         & \multicolumn{1}{c|}{0.4660}                         & \multicolumn{1}{c|}{{0.5585}}                   & \multicolumn{1}{c|}{{0.5592}}                   & \multicolumn{1}{c|}{{0.0858}}                   & {0.0865}                   \\
\multicolumn{1}{c|}{}                     & \multicolumn{1}{c|}{MIST}                          & \multicolumn{1}{c|}{{17.32}}                   & \multicolumn{1}{c|}{17.32}                         & \multicolumn{1}{c|}{88.17}                          & \multicolumn{1}{c|}{88.87}                          & \multicolumn{1}{c|}{0.4616}                         & \multicolumn{1}{c|}{0.4632}                         & \multicolumn{1}{c|}{0.5539}                         & \multicolumn{1}{c|}{0.5552}                         & \multicolumn{1}{c|}{0.0830}                         & 0.0837                         \\
\multicolumn{1}{c|}{}                     & \multicolumn{1}{c|}{Ours-}                         & \multicolumn{1}{c|}{\textbf{16.74}}                & \multicolumn{1}{c|}{\textbf{16.70}}                & \multicolumn{1}{c|}{70.59}                          & \multicolumn{1}{c|}{69.04}                          & \multicolumn{1}{c|}{\textbf{0.4336}}                         & \multicolumn{1}{c|}{\textbf{0.4329}}                         & \multicolumn{1}{c|}{\textbf{0.6005}}                & \multicolumn{1}{c|}{\textbf{0.5990}}                & \multicolumn{1}{c|}{\textbf{0.1156}}                & \textbf{0.1167}                \\ \hline
\multicolumn{1}{c|}{\multirow{3}{*}{P3}}     & \multicolumn{1}{c|}{PG}                            & \multicolumn{1}{c|}{{17.26}}                   & \multicolumn{1}{c|}{17.22}                         & \multicolumn{1}{c|}{123.94}                         & \multicolumn{1}{c|}{127.87}                         & \multicolumn{1}{c|}{0.4702}                         & \multicolumn{1}{c|}{0.4749}                         & \multicolumn{1}{c|}{{0.5727}}                   & \multicolumn{1}{c|}{{0.5727}}                   & \multicolumn{1}{c|}{{0.0847}}                   & {0.0853}                   \\
\multicolumn{1}{c|}{}                     & \multicolumn{1}{c|}{MIST}                          & \multicolumn{1}{c|}{17.27}                         & \multicolumn{1}{c|}{{17.18}}                   & \multicolumn{1}{c|}{121.36}                         & \multicolumn{1}{c|}{127.34}                         & \multicolumn{1}{c|}{0.4685}                         & \multicolumn{1}{c|}{0.4732}                         & \multicolumn{1}{c|}{0.5691}                         & \multicolumn{1}{c|}{0.5687}                         & \multicolumn{1}{c|}{0.0820}                         & 0.0829                         \\
\multicolumn{1}{c|}{}                     & \multicolumn{1}{c|}{Ours-}                & \multicolumn{1}{c|}{\textbf{16.71}}                & \multicolumn{1}{c|}{\textbf{16.78}}                & \multicolumn{1}{c|}{\textbf{156.47}}                & \multicolumn{1}{c|}{\textbf{153.85}}                & \multicolumn{1}{c|}{\textbf{0.4234}}                & \multicolumn{1}{c|}{\textbf{0.4228}}                & \multicolumn{1}{c|}{\textbf{0.6202}}                & \multicolumn{1}{c|}{\textbf{0.6204}}                & \multicolumn{1}{c|}{\textbf{0.1084}}                & \textbf{0.1079}                \\\hline
\multicolumn{1}{c|}{\multirow{3}{*}{P4}}      & \multicolumn{1}{c|}{PG}                            & \multicolumn{1}{c|}{{17.87}}                   & \multicolumn{1}{c|}{{17.84}}                   & \multicolumn{1}{c|}{\textbf{150.85}}                & \multicolumn{1}{c|}{\textbf{151.75}}                & \multicolumn{1}{c|}{0.4560}                         & \multicolumn{1}{c|}{0.4589}                         & \multicolumn{1}{c|}{{0.5820}}                   & \multicolumn{1}{c|}{{0.5818}}                   & \multicolumn{1}{c|}{{0.0804}}                   & {0.0819}                   \\
\multicolumn{1}{c|}{}                     & \multicolumn{1}{c|}{MIST}                          & \multicolumn{1}{c|}{17.88}                         & \multicolumn{1}{c|}{17.85}                         & \multicolumn{1}{c|}{148.64}                         & \multicolumn{1}{c|}{{150.45}}                   & \multicolumn{1}{c|}{0.4526}                         & \multicolumn{1}{c|}{0.4540}                         & \multicolumn{1}{c|}{0.5786}                         & \multicolumn{1}{c|}{0.5788}                         & \multicolumn{1}{c|}{0.0770}                         & 0.0786                         \\
\multicolumn{1}{c|}{}                     & \multicolumn{1}{c|}{Ours-}                         & \multicolumn{1}{c|}{\textbf{17.61}}                & \multicolumn{1}{c|}{\textbf{17.50}}                & \multicolumn{1}{c|}{116.48}                         & \multicolumn{1}{c|}{115.41}                         & \multicolumn{1}{c|}{\textbf{0.4004}}                & \multicolumn{1}{c|}{\textbf{0.3942}}                & \multicolumn{1}{c|}{\textbf{0.6361}}                & \multicolumn{1}{c|}{\textbf{0.6394}}                & \multicolumn{1}{c|}{\textbf{0.1051}}                & \textbf{0.1067}                \\ \hline
\multicolumn{1}{c|}{\multirow{3}{*}{P5}}       & \multicolumn{1}{c|}{PG}                            & \multicolumn{1}{c|}{{17.77}}                   & \multicolumn{1}{c|}{{17.80}}                   & \multicolumn{1}{c|}{110.96}                         & \multicolumn{1}{c|}{112.49}                         & \multicolumn{1}{c|}{0.5091}                         & \multicolumn{1}{c|}{0.4988}                         & \multicolumn{1}{c|}{{0.5460}}                   & \multicolumn{1}{c|}{{0.5517}}                   & \multicolumn{1}{c|}{{0.0945}}                   & {0.0924}                   \\
\multicolumn{1}{c|}{}                     & \multicolumn{1}{c|}{MIST}                          & \multicolumn{1}{c|}{17.81}                         & \multicolumn{1}{c|}{17.82}                         & \multicolumn{1}{c|}{107.88}                         & \multicolumn{1}{c|}{110.79}                         & \multicolumn{1}{c|}{0.5087}                         & \multicolumn{1}{c|}{0.4972}                         & \multicolumn{1}{c|}{0.5404}                         & \multicolumn{1}{c|}{0.5459}                         & \multicolumn{1}{c|}{0.0915}                         & 0.0892                         \\
\multicolumn{1}{c|}{}                     & \multicolumn{1}{c|}{Ours-}                         & \multicolumn{1}{c|}{\textbf{16.81}}                & \multicolumn{1}{c|}{\textbf{16.95}}                & \multicolumn{1}{c|}{\textbf{134.71}}                         & \multicolumn{1}{c|}{\textbf{129.93}}                         & \multicolumn{1}{c|}{\textbf{0.4568}}                         & \multicolumn{1}{c|}{\textbf{0.4524}}                         & \multicolumn{1}{c|}{\textbf{0.6032}}                & \multicolumn{1}{c|}{\textbf{0.6047}}                & \multicolumn{1}{c|}{\textbf{0.1227}}                & \textbf{0.1209}                \\\hline
\end{tabular}
\end{table*}

%% file: float/extra-prompts-obj1.tex
\begin{table}[t]
\centering
\caption{Results on Claude-generated prompt editing for methods subject to Objective 1.}
\label{tab:extended:obj1}
\begin{tabular}{c|ccccc}\hline
      & PSNR$\uparrow$ & FID$\downarrow$ & SSIM$\uparrow$ & LPIPS$\downarrow$ & ACDM$\downarrow$ \\\hline
clean & 15.35          & 153.05          & 0.6125         & 0.4077            & 0.0515           \\\hline
AdvDM & 17.43          & 144.61          & 0.2719         & 0.6267            & \textbf{0.0433}  \\
SDS   & 17.36          & 146.43          & 0.2701         & 0.6300            & 0.0496           \\
PCA+  & \textbf{19.06} & \textbf{109.65} & \textbf{0.3910}& \textbf{0.6065}   & 0.0439           \\\hline    
\end{tabular}
\end{table}

%% file: float/extra-prompts-obj2.tex
\begin{table}[t]
\centering
\caption{Results on Claude-generated prompt editing for methods subject to Objective 2.}
\label{tab:extended:obj2}
\begin{tabular}{c|ccccc}\hline
     & PSNR$\downarrow$ & FID$\uparrow$   & SSIM$\downarrow$ & LPIPS$\uparrow$ & ACDM$\uparrow$  \\\hline
PG   & 14.26            & 181.93          & 0.3973           & 0.6308          & 0.1041          \\
MIST & 14.17            & \textbf{194.50} & 0.3543           & 0.6213          & 0.0953          \\
PCA- & \textbf{13.92}   & 166.32          & \textbf{0.3276}  & \textbf{0.6468} & \textbf{0.1413} \\\hline
\end{tabular}
\end{table}

%% file: float/defense-adv-clean.tex
\begin{table}[t]
\centering
\caption{Defense against Adv-Clean, which evaluates the effectiveness after applying image degradation and filtering operations.}\label{tab:defense:advclean}
\resizebox{\linewidth}{!}{
\begin{tabular}{c|c|ccc}\hline
    Objective 1 & clean & AdvDM & SDS & PCA+ \\\hline
    PSNR$\uparrow$ & $18.74_{\pm0.23}$ & $17.56_{\pm0.22}$ & $17.70_{\pm0.20}$ & $\mathbf{18.94_{\pm0.21}}$ \\
    FID$\downarrow$ & $66.04_{\pm15.87}$ & $107.18_{\pm28.29}$ & $109.14_{\pm26.56}$ & $\mathbf{92.65_{\pm21.43}}$\\
    SSIM$\uparrow$ & $0.5826_{\pm0.0087}$ & $0.4349_{\pm0.0086}$ & $0.4490_{\pm0.0086}$ & $\mathbf{0.5177_{\pm0.0100}}$\\
    LPIPS$\downarrow$ & $0.4121_{\pm0.0164}$ & $0.5029_{\pm0.0136}$ & $0.5008_{\pm0.0125}$ & $\mathbf{0.4953_{\pm0.0094}}$\\
    ACDM$\downarrow$ & $0.0257_{\pm0.0010}$ & $0.0395_{\pm0.0013}$ & $0.0410_{\pm0.0013}$ & $\mathbf{0.0458_{\pm0.0010}}$\\ \hline
\end{tabular}
}

\vspace{1em}
\begin{tabular}{c|ccc}\hline
    Objective 2 & PhotoGuard & MIST & PCA- \\\hline
    PSNR$\downarrow$ & $19.74_{\pm0.71}$ & $19.76_{\pm0.71}$ & $\mathbf{18.88_{\pm0.69}}$ \\
    FID$\uparrow$ & $83.88_{\pm18.53}$ & $84.21_{\pm18.82}$ & $\mathbf{100.25_{\pm24.07}}$ \\
    SSIM$\downarrow$ & $0.6019_{\pm0.0168}$ & $0.6053_{\pm0.0168}$ & $\mathbf{0.5224_{\pm0.0179}}$ \\
    LPIPS$\uparrow$ & $0.4824_{\pm0.0216}$ & $0.4770_{\pm0.0223}$ & $\mathbf{0.5462_{\pm0.0187}}$ \\
    ACDM$\uparrow$ & $0.0572_{\pm0.0043}$ & $0.0558_{\pm0.0043}$ & $\mathbf{0.0723_{\pm0.0068}}$ \\ \hline
\end{tabular}

\vspace{0.5em}
\emph{(Values are reported as} $\text{mean}_{\pm\text{std}}$ \emph{across different prompts.)}
\end{table}

%% file: float/defense-diffpure.tex
\begin{table}[t]
\centering
\caption{Defense against DiffPure, which evaluates the effectiveness against Diffusion-based adversarial purification.}\label{tab:defense:diffpure}
\resizebox{\linewidth}{!}{
\begin{tabular}{c|c|ccc}\hline
Objective 1 & clean & AdvDM & SDS & Ours+ \\\hline
PSNR$\uparrow$ & $18.74_{\pm 0.23}$ & $18.88_{\pm 0.21}$ & $18.87_{\pm 0.21}$ & $\mathbf{19.16_{\pm 0.20}}$ \\
FID$\downarrow$ & $66.04_{\pm 15.87}$ & $69.90_{\pm 16.88}$ & $\mathbf{68.94_{\pm 16.32}}$ & $71.12_{\pm 16.40}$ \\
SSIM$\uparrow$ & $0.5826_{\pm 0.0087}$ & $0.5683_{\pm 0.0098}$ & $0.5689_{\pm 0.0098}$ & $\mathbf{0.5731_{\pm 0.0088}}$ \\
LPIPS$\downarrow$ & $0.4121_{\pm 0.0164}$ & $\mathbf{0.4349_{\pm 0.0152}}$ & $0.4370_{\pm 0.0148}$ & $0.4437_{\pm 0.0142}$ \\
ACDM$\downarrow$ & $0.0257_{\pm 0.0010}$ & $\mathbf{0.0298_{\pm 0.0010}}$ & $0.0301_{\pm 0.0010}$ & $0.0351_{\pm 0.0010}$ \\
\hline
\end{tabular}
}

\vspace{1em}
\begin{tabular}{c|ccc}
\hline
Objective 2 & PhotoGuard & MIST & PCA- \\
\hline
PSNR$\downarrow$ & $24.45_{\pm 1.05}$ & $24.49_{\pm 1.05}$ & $\mathbf{24.21_{\pm 1.02}}$ \\
FID$\uparrow$ & $40.35_{\pm 6.42}$ & $39.84_{\pm 6.52}$ & $\mathbf{43.45_{\pm 6.17}}$ \\
SSIM$\downarrow$ & $0.8002_{\pm 0.0184}$ & $0.8014_{\pm 0.0184}$ & $\mathbf{0.7868_{\pm 0.0185}}$ \\
LPIPS$\uparrow$ & $0.2669_{\pm 0.0239}$ & $0.2643_{\pm 0.0237}$ & $\mathbf{0.2804_{\pm 0.0245}}$ \\
ACDM$\uparrow$ & $0.0246_{\pm 0.0009}$ & $0.0240_{\pm 0.0008}$ & $\mathbf{0.0296_{\pm 0.0010}}$ \\
\hline
\end{tabular}

\vspace{0.5em}
\emph{(Values are reported as} $\text{mean}_{\pm\text{std}}$ \emph{across different prompts.)}
\end{table}

%% file: float/ablation-v-objs.tex
\begin{table*}[ht]
\centering
\caption{Hyperparameters analysis on $v$ for PCA+.}
\begin{tabular}{cccccccccccc}
\hline
\multirow{2}{*}{prompt}                  & \multirow{2}{*}{$v$}         & \multicolumn{2}{c}{PSNR$\uparrow$}                      & \multicolumn{2}{c}{FID$\downarrow$}                       & \multicolumn{2}{c}{SSIM$\uparrow$}                        & \multicolumn{2}{c}{LPIPS$\downarrow$}                     & \multicolumn{2}{c}{ACDM$\downarrow$} \\
                                         &                            & SD14                       & SD15                       & SD14                        & SD15                        & SD14                        & SD15                        & SD14                        & SD15                        & SD14                        & SD15   \\ \hline
\multicolumn{1}{c|}{\multirow{3}{*}{P1}} & \multicolumn{1}{c|}{clean} & 19.09                      & 19.09                      & 54.95                       & 54.25                       & 0.5911                      & 0.5914                      & 0.3948                      & 0.3943                      & 0.0243                      & 0.0242 \\ \cline{2-12} 
\multicolumn{1}{c|}{}                    & \multicolumn{1}{c|}{1}     & \multicolumn{1}{c|}{\textbf{18.85}} & \multicolumn{1}{c|}{\textbf{18.82}} & \multicolumn{1}{c|}{\textbf{109.01}} & \multicolumn{1}{c|}{\textbf{108.57}} & \multicolumn{1}{c|}{\textbf{0.4424}} & \multicolumn{1}{c|}{\textbf{0.4396}} & \multicolumn{1}{c|}{\textbf{0.5349}} & \multicolumn{1}{c|}{\textbf{0.5345}} & \multicolumn{1}{c|}{\textbf{0.0593}} & \textbf{0.0601} \\
\multicolumn{1}{c|}{}                    & \multicolumn{1}{c|}{1e-8}  & \multicolumn{1}{c|}{18.16} & \multicolumn{1}{c|}{18.13} & \multicolumn{1}{c|}{112.57} & \multicolumn{1}{c|}{111.86} & \multicolumn{1}{c|}{0.4212} & \multicolumn{1}{c|}{0.4275} & \multicolumn{1}{c|}{0.5486} & \multicolumn{1}{c|}{0.5485} & \multicolumn{1}{c|}{0.0602} & 0.0610 \\ \hline
\multicolumn{1}{c|}{\multirow{3}{*}{P2}} & \multicolumn{1}{c|}{clean} & 18.68                      & 18.69                      & 45.34                       & 45.41                       & 0.5829                      & 0.5838                      & 0.3993                      & 0.3986                      & 0.0253                      & 0.0251 \\ \cline{2-12} 
\multicolumn{1}{c|}{}                    & \multicolumn{1}{c|}{1}     & \multicolumn{1}{c|}{\textbf{18.61}} & \multicolumn{1}{c|}{\textbf{18.60}} & \multicolumn{1}{c|}{\textbf{70.40}}  & \multicolumn{1}{c|}{\textbf{70.08}}  & \multicolumn{1}{c|}{\textbf{0.4460}} & \multicolumn{1}{c|}{\textbf{0.4444}} & \multicolumn{1}{c|}{\textbf{0.5276}} & \multicolumn{1}{c|}{\textbf{0.5269}} & \multicolumn{1}{c|}{\textbf{0.0602}} & \textbf{0.0608} \\
\multicolumn{1}{c|}{}                    & \multicolumn{1}{c|}{1e-8}  & \multicolumn{1}{c|}{18.47} & \multicolumn{1}{c|}{18.50} & \multicolumn{1}{c|}{72.28}  & \multicolumn{1}{c|}{73.06}  & \multicolumn{1}{c|}{0.4363} & \multicolumn{1}{c|}{0.4335} & \multicolumn{1}{c|}{0.5302} & \multicolumn{1}{c|}{0.5304} & \multicolumn{1}{c|}{0.0610} & 0.0616 \\ \hline
\multicolumn{1}{c|}{\multirow{3}{*}{P3}} & \multicolumn{1}{c|}{clean} & 18.50                      & 18.54                      & 78.89                       & 78.98                       & 0.5722                      & 0.5735                      & 0.4284                      & 0.4281                      & 0.0268                      & 0.0266 \\ \cline{2-12} 
\multicolumn{1}{c|}{}                    & \multicolumn{1}{c|}{1}     & \multicolumn{1}{c|}{\textbf{18.55}} & \multicolumn{1}{c|}{\textbf{18.58}} & \multicolumn{1}{c|}{\textbf{114.34}} & \multicolumn{1}{c|}{\textbf{112.30}} & \multicolumn{1}{c|}{\textbf{0.4348}} & \multicolumn{1}{c|}{\textbf{0.4355}} & \multicolumn{1}{c|}{\textbf{0.5420}} & \multicolumn{1}{c|}{\textbf{0.5405}} & \multicolumn{1}{c|}{\textbf{0.0605}} & \textbf{0.0605} \\
\multicolumn{1}{c|}{}                    & \multicolumn{1}{c|}{1e-8}  & \multicolumn{1}{c|}{18.49} & \multicolumn{1}{c|}{18.54} & \multicolumn{1}{c|}{116.97} & \multicolumn{1}{c|}{116.55} & \multicolumn{1}{c|}{0.4242} & \multicolumn{1}{c|}{0.4237} & \multicolumn{1}{c|}{0.5548} & \multicolumn{1}{c|}{0.5534} & \multicolumn{1}{c|}{0.0611} & 0.0612 \\ \hline
\multicolumn{1}{c|}{\multirow{3}{*}{P4}} & \multicolumn{1}{c|}{clean} & 18.82                      & 18.84                      & 69.26                       & 69.17                       & 0.5886                      & 0.5905                      & 0.4116                      & 0.4096                      & 0.0256                      & 0.0256 \\ \cline{2-12} 
\multicolumn{1}{c|}{}                    & \multicolumn{1}{c|}{1}     & \multicolumn{1}{c|}{\textbf{18.72}} & \multicolumn{1}{c|}{\textbf{18.77}} & \multicolumn{1}{c|}{\textbf{110.03}} & \multicolumn{1}{c|}{\textbf{109.44}} & \multicolumn{1}{c|}{\textbf{0.4447}} & \multicolumn{1}{c|}{\textbf{0.4447}} & \multicolumn{1}{c|}{\textbf{0.5396}} & \multicolumn{1}{c|}{\textbf{0.5385}} & \multicolumn{1}{c|}{\textbf{0.0614}} & \textbf{0.0609} \\
\multicolumn{1}{c|}{}                    & \multicolumn{1}{c|}{1e-8}  & \multicolumn{1}{c|}{18.70} & \multicolumn{1}{c|}{18.76} & \multicolumn{1}{c|}{114.24} & \multicolumn{1}{c|}{112.44} & \multicolumn{1}{c|}{0.4339} & \multicolumn{1}{c|}{0.4323} & \multicolumn{1}{c|}{0.5429} & \multicolumn{1}{c|}{0.5424} & \multicolumn{1}{c|}{0.0616} & 0.0613 \\ \hline
\multicolumn{1}{c|}{\multirow{3}{*}{P5}} & \multicolumn{1}{c|}{clean} & 18.52                      & 18.55                      & 83.84                       & 82.33                       & 0.5724                      & 0.5737                      & 0.4321                      & 0.4298                      & 0.0268                      & 0.0267 \\ \cline{2-12} 
\multicolumn{1}{c|}{}                    & \multicolumn{1}{c|}{1}     & \multicolumn{1}{c|}{\textbf{18.50}} & \multicolumn{1}{c|}{\textbf{18.40}} & \multicolumn{1}{c|}{\textbf{124.06}} & \multicolumn{1}{c|}{\textbf{127.72}} & \multicolumn{1}{c|}{\textbf{0.4144}} & \multicolumn{1}{c|}{\textbf{0.4092}} & \multicolumn{1}{c|}{\textbf{0.5441}} & \multicolumn{1}{c|}{\textbf{0.5444}} & \multicolumn{1}{c|}{\textbf{0.0618}} & \textbf{0.0640} \\
\multicolumn{1}{c|}{}                    & \multicolumn{1}{c|}{1e-8}  & \multicolumn{1}{c|}{18.45} & \multicolumn{1}{c|}{18.39} & \multicolumn{1}{c|}{129.09} & \multicolumn{1}{c|}{130.72} & \multicolumn{1}{c|}{0.4033} & \multicolumn{1}{c|}{0.4073} & \multicolumn{1}{c|}{0.5495} & \multicolumn{1}{c|}{0.5501} & \multicolumn{1}{c|}{0.0622} & 0.0652 \\ \hline
\end{tabular}
\label{tab:abl:v:obj1}
\end{table*}

%% file: float/ablation-v.tex
\begin{table*}[ht]
\centering
\caption{Hyperparameters analysis on $v$ for PCA-.}
\begin{tabular}{ccllllllllll}
\hline
\multirow{2}{*}{prompt}                  & \multirow{2}{*}{$v$}      & \multicolumn{2}{c}{PSNR$\downarrow$}                                        & \multicolumn{2}{c}{FID$\uparrow$}                                         & \multicolumn{2}{c}{SSIM$\downarrow$}                                          & \multicolumn{2}{c}{LPIPS$\uparrow$}                                       & \multicolumn{2}{c}{ACDM$\uparrow$}                            \\
                                         &                           & \multicolumn{1}{c}{SD14}            & \multicolumn{1}{c}{SD15}            & \multicolumn{1}{c}{SD14}             & \multicolumn{1}{c}{SD15}             & \multicolumn{1}{c}{SD14}             & \multicolumn{1}{c}{SD15}             & \multicolumn{1}{c}{SD14}             & \multicolumn{1}{c}{SD15}             & \multicolumn{1}{c}{SD14}             & \multicolumn{1}{c}{SD15} \\ \hline
\multicolumn{1}{c|}{\multirow{2}{*}{P1}} & \multicolumn{1}{c|}{1}    & \multicolumn{1}{c|}{18.65}          & \multicolumn{1}{c|}{18.65}          & \multicolumn{1}{c|}{124.33}          & \multicolumn{1}{c|}{124.37}          & \multicolumn{1}{c|}{0.4299}          & \multicolumn{1}{c|}{0.4317}          & \multicolumn{1}{c|}{0.5932}          & \multicolumn{1}{c|}{0.5927}          & \multicolumn{1}{c|}{0.0838}          & 0.0844                   \\
\multicolumn{1}{c|}{}                    & \multicolumn{1}{c|}{1e-8} & \multicolumn{1}{c|}{\textbf{18.02}} & \multicolumn{1}{c|}{\textbf{18.02}} & \multicolumn{1}{c|}{\textbf{139.11}} & \multicolumn{1}{c|}{\textbf{135.32}} & \multicolumn{1}{c|}{\textbf{0.4441}} & \multicolumn{1}{c|}{\textbf{0.4453}} & \multicolumn{1}{c|}{\textbf{0.6188}} & \multicolumn{1}{c|}{\textbf{0.6171}} & \multicolumn{1}{c|}{\textbf{0.1106}} & \textbf{0.1110}          \\ \hline
\multicolumn{1}{c|}{\multirow{2}{*}{P2}} & \multicolumn{1}{c|}{1}    & \multicolumn{1}{c|}{17.34}          & \multicolumn{1}{c|}{17.31}          & \multicolumn{1}{c|}{63.33}           & \multicolumn{1}{c|}{62.23}           & \multicolumn{1}{c|}{0.4251}          & \multicolumn{1}{c|}{0.4244}          & \multicolumn{1}{c|}{0.5768}          & \multicolumn{1}{c|}{0.5758}          & \multicolumn{1}{c|}{0.0897}          & 0.0898                   \\
\multicolumn{1}{c|}{}                    & \multicolumn{1}{c|}{1e-8} & \multicolumn{1}{c|}{\textbf{16.74}} & \multicolumn{1}{c|}{\textbf{16.70}} & \multicolumn{1}{c|}{\textbf{70.59}}  & \multicolumn{1}{c|}{\textbf{69.04}}  & \multicolumn{1}{c|}{\textbf{0.4336}} & \multicolumn{1}{c|}{\textbf{0.4329}} & \multicolumn{1}{c|}{\textbf{0.6005}} & \multicolumn{1}{c|}{\textbf{0.5990}} & \multicolumn{1}{c|}{\textbf{0.1156}} & \textbf{0.1167}          \\ \hline
\multicolumn{1}{c|}{\multirow{2}{*}{P3}} & \multicolumn{1}{c|}{1}    & \multicolumn{1}{c|}{17.12}          & \multicolumn{1}{c|}{17.15}          & \multicolumn{1}{c|}{128.91}          & \multicolumn{1}{c|}{128.16}          & \multicolumn{1}{c|}{0.4108}          & \multicolumn{1}{c|}{0.4100}          & \multicolumn{1}{c|}{0.6024}          & \multicolumn{1}{c|}{0.6027}          & \multicolumn{1}{c|}{0.0856}          & 0.0851                   \\
\multicolumn{1}{c|}{}                    & \multicolumn{1}{c|}{1e-8} & \multicolumn{1}{c|}{\textbf{16.71}} & \multicolumn{1}{c|}{\textbf{16.78}} & \multicolumn{1}{c|}{\textbf{156.47}} & \multicolumn{1}{c|}{\textbf{153.85}} & \multicolumn{1}{c|}{\textbf{0.4234}} & \multicolumn{1}{c|}{\textbf{0.4228}} & \multicolumn{1}{c|}{\textbf{0.6202}} & \multicolumn{1}{c|}{\textbf{0.6204}} & \multicolumn{1}{c|}{\textbf{0.1084}} & \textbf{0.1079}          \\ \hline
\multicolumn{1}{c|}{\multirow{2}{*}{P4}} & \multicolumn{1}{c|}{1}    & \multicolumn{1}{c|}{18.08}          & \multicolumn{1}{c|}{18.00}          & \multicolumn{1}{c|}{106.21}          & \multicolumn{1}{c|}{103.85}          & \multicolumn{1}{c|}{0.3768}          & \multicolumn{1}{c|}{0.3745}          & \multicolumn{1}{c|}{0.6305}          & \multicolumn{1}{c|}{0.6325}          & \multicolumn{1}{c|}{0.0784}          & 0.0787                   \\
\multicolumn{1}{c|}{}                    & \multicolumn{1}{c|}{1e-8} & \multicolumn{1}{c|}{\textbf{17.61}} & \multicolumn{1}{c|}{\textbf{17.50}} & \multicolumn{1}{c|}{\textbf{116.48}} & \multicolumn{1}{c|}{\textbf{115.41}} & \multicolumn{1}{c|}{\textbf{0.4004}} & \multicolumn{1}{c|}{\textbf{0.3942}} & \multicolumn{1}{c|}{\textbf{0.6361}} & \multicolumn{1}{c|}{\textbf{0.6394}} & \multicolumn{1}{c|}{\textbf{0.1051}} & \textbf{0.1067}          \\ \hline
\multicolumn{1}{c|}{\multirow{2}{*}{P5}} & \multicolumn{1}{c|}{1}    & \multicolumn{1}{c|}{17.44}          & \multicolumn{1}{c|}{17.59}          & \multicolumn{1}{c|}{120.47}          & \multicolumn{1}{c|}{116.90}          & \multicolumn{1}{c|}{0.4491}          & \multicolumn{1}{c|}{0.4436}          & \multicolumn{1}{c|}{0.5871}          & \multicolumn{1}{c|}{0.5888}          & \multicolumn{1}{c|}{0.0957}          & 0.0939                   \\
\multicolumn{1}{c|}{}                    & \multicolumn{1}{c|}{1e-8} & \multicolumn{1}{c|}{\textbf{16.81}} & \multicolumn{1}{c|}{\textbf{16.95}} & \multicolumn{1}{c|}{\textbf{134.71}} & \multicolumn{1}{c|}{\textbf{129.93}} & \multicolumn{1}{c|}{\textbf{0.4568}} & \multicolumn{1}{c|}{\textbf{0.4524}} & \multicolumn{1}{c|}{\textbf{0.6032}} & \multicolumn{1}{c|}{\textbf{0.6047}} & \multicolumn{1}{c|}{\textbf{0.1227}} & \textbf{0.1209}          \\ \hline
\end{tabular}
\label{tab:ablation:v}
\end{table*}

%% file: float/ablation-alpha.tex
\begin{table*}[t]
\centering
\caption{Hyperparameters analysis on $\alpha$.}
\begin{tabular}{cccccccccccc}
\hline
\multirow{2}{*}{prompt}                  & \multirow{2}{*}{$\alpha$} & \multicolumn{2}{c}{PSNR$\uparrow$}                                        & \multicolumn{2}{c}{FID$\downarrow$}                                         & \multicolumn{2}{c}{SSIM$\uparrow$}                                          & \multicolumn{2}{c}{LPIPS$\downarrow$}                                       & \multicolumn{2}{c}{ACDM$\downarrow$}                   \\
                                         &                           & SD14                                & SD15                                & SD14                                 & SD15                                 & SD14                                 & SD15                                 & SD14                                 & SD15                                 & SD14                                 & SD15            \\ \hline
\multicolumn{1}{c|}{\multirow{3}{*}{P1}} & \multicolumn{1}{c|}{1}    & \multicolumn{1}{c|}{18.18}          & \multicolumn{1}{c|}{18.18}          & \multicolumn{1}{c|}{137.44}          & \multicolumn{1}{c|}{134.12}          & \multicolumn{1}{c|}{0.4563}          & \multicolumn{1}{c|}{0.4576}          & \multicolumn{1}{c|}{0.6137}          & \multicolumn{1}{c|}{0.6119}          & \multicolumn{1}{c|}{0.1075}          & 0.1080          \\
\multicolumn{1}{c|}{}                    & \multicolumn{1}{c|}{2}    & \multicolumn{1}{c|}{\textbf{18.02}} & \multicolumn{1}{c|}{\textbf{18.02}} & \multicolumn{1}{c|}{\textbf{139.11}} & \multicolumn{1}{c|}{\textbf{135.32}} & \multicolumn{1}{c|}{0.4441}          & \multicolumn{1}{c|}{0.4453}          & \multicolumn{1}{c|}{\textbf{0.6188}} & \multicolumn{1}{c|}{\textbf{0.6171}} & \multicolumn{1}{c|}{\textbf{0.1106}} & \textbf{0.1110} \\
\multicolumn{1}{c|}{}                    & \multicolumn{1}{c|}{4}    & \multicolumn{1}{c|}{18.43}          & \multicolumn{1}{c|}{18.44}          & \multicolumn{1}{c|}{132.79}          & \multicolumn{1}{c|}{132.77}          & \multicolumn{1}{c|}{\textbf{0.4408}} & \multicolumn{1}{c|}{\textbf{0.4419}} & \multicolumn{1}{c|}{0.6057}          & \multicolumn{1}{c|}{0.6049}          & \multicolumn{1}{c|}{0.0961}          & 0.0966          \\ \hline
\multicolumn{1}{c|}{\multirow{3}{*}{P2}} & \multicolumn{1}{c|}{1}    & \multicolumn{1}{c|}{16.85}          & \multicolumn{1}{c|}{16.82}          & \multicolumn{1}{c|}{70.40}           & \multicolumn{1}{c|}{68.52}           & \multicolumn{1}{c|}{0.4448}          & \multicolumn{1}{c|}{0.4445}          & \multicolumn{1}{c|}{0.5956}          & \multicolumn{1}{c|}{0.5938}          & \multicolumn{1}{c|}{0.1131}          & 0.1142          \\
\multicolumn{1}{c|}{}                    & \multicolumn{1}{c|}{2}    & \multicolumn{1}{c|}{\textbf{16.74}} & \multicolumn{1}{c|}{\textbf{16.70}} & \multicolumn{1}{c|}{\textbf{70.59}}  & \multicolumn{1}{c|}{\textbf{69.04}}  & \multicolumn{1}{c|}{0.4336}          & \multicolumn{1}{c|}{0.4329}          & \multicolumn{1}{c|}{\textbf{0.6005}} & \multicolumn{1}{c|}{\textbf{0.5990}} & \multicolumn{1}{c|}{\textbf{0.1156}} & \textbf{0.1167} \\
\multicolumn{1}{c|}{}                    & \multicolumn{1}{c|}{4}    & \multicolumn{1}{c|}{17.10}          & \multicolumn{1}{c|}{17.08}          & \multicolumn{1}{c|}{66.88}           & \multicolumn{1}{c|}{65.54}           & \multicolumn{1}{c|}{\textbf{0.4330}} & \multicolumn{1}{c|}{\textbf{0.4327}} & \multicolumn{1}{c|}{0.5887}          & \multicolumn{1}{c|}{0.5863}          & \multicolumn{1}{c|}{0.1018}          & 0.1026          \\ \hline
\multicolumn{1}{c|}{\multirow{3}{*}{P3}} & \multicolumn{1}{c|}{1}    & \multicolumn{1}{c|}{16.86}          & \multicolumn{1}{c|}{16.91}          & \multicolumn{1}{c|}{155.70}          & \multicolumn{1}{c|}{152.69}          & \multicolumn{1}{c|}{0.4330}          & \multicolumn{1}{c|}{0.4323}          & \multicolumn{1}{c|}{0.6165}          & \multicolumn{1}{c|}{0.6171}          & \multicolumn{1}{c|}{0.1052}          & 0.1053          \\
\multicolumn{1}{c|}{}                    & \multicolumn{1}{c|}{2}    & \multicolumn{1}{c|}{\textbf{16.71}} & \multicolumn{1}{c|}{\textbf{16.78}} & \multicolumn{1}{c|}{\textbf{156.47}} & \multicolumn{1}{c|}{\textbf{153.85}} & \multicolumn{1}{c|}{0.4234}          & \multicolumn{1}{c|}{0.4228}          & \multicolumn{1}{c|}{\textbf{0.6202}} & \multicolumn{1}{c|}{\textbf{0.6204}} & \multicolumn{1}{c|}{\textbf{0.1084}} & \textbf{0.1079} \\
\multicolumn{1}{c|}{}                    & \multicolumn{1}{c|}{4}    & \multicolumn{1}{c|}{17.00}          & \multicolumn{1}{c|}{17.04}          & \multicolumn{1}{c|}{140.65}          & \multicolumn{1}{c|}{139.28}          & \multicolumn{1}{c|}{\textbf{0.4198}} & \multicolumn{1}{c|}{\textbf{0.4189}} & \multicolumn{1}{c|}{0.6110}          & \multicolumn{1}{c|}{0.6108}          & \multicolumn{1}{c|}{0.0958}          & 0.0956          \\ \hline
\multicolumn{1}{c|}{\multirow{3}{*}{P4}} & \multicolumn{1}{c|}{1}    & \multicolumn{1}{c|}{17.75}          & \multicolumn{1}{c|}{17.62}          & \multicolumn{1}{c|}{\textbf{118.34}} & \multicolumn{1}{c|}{113.48}          & \multicolumn{1}{c|}{0.4138}          & \multicolumn{1}{c|}{0.4067}          & \multicolumn{1}{c|}{0.6305}          & \multicolumn{1}{c|}{0.6337}          & \multicolumn{1}{c|}{0.1021}          & 0.1036          \\
\multicolumn{1}{c|}{}                    & \multicolumn{1}{c|}{2}    & \multicolumn{1}{c|}{\textbf{17.61}} & \multicolumn{1}{c|}{\textbf{17.50}} & \multicolumn{1}{c|}{116.48}          & \multicolumn{1}{c|}{\textbf{115.41}} & \multicolumn{1}{c|}{0.4004}          & \multicolumn{1}{c|}{0.3942}          & \multicolumn{1}{c|}{\textbf{0.6361}} & \multicolumn{1}{c|}{\textbf{0.6394}} & \multicolumn{1}{c|}{\textbf{0.1051}} & \textbf{0.1067} \\
\multicolumn{1}{c|}{}                    & \multicolumn{1}{c|}{4}    & \multicolumn{1}{c|}{17.96}          & \multicolumn{1}{c|}{17.85}          & \multicolumn{1}{c|}{110.58}          & \multicolumn{1}{c|}{109.59}          & \multicolumn{1}{c|}{\textbf{0.3929}} & \multicolumn{1}{c|}{\textbf{0.3876}} & \multicolumn{1}{c|}{0.6302}          & \multicolumn{1}{c|}{0.6347}          & \multicolumn{1}{c|}{0.0918}          & 0.0927          \\ \hline
\multicolumn{1}{c|}{\multirow{3}{*}{P5}} & \multicolumn{1}{c|}{1}    & \multicolumn{1}{c|}{16.91}          & \multicolumn{1}{c|}{17.06}          & \multicolumn{1}{c|}{132.37}          & \multicolumn{1}{c|}{128.71}          & \multicolumn{1}{c|}{0.4670}          & \multicolumn{1}{c|}{0.4628}          & \multicolumn{1}{c|}{0.5989}          & \multicolumn{1}{c|}{0.6004}          & \multicolumn{1}{c|}{0.1207}          & 0.1182          \\
\multicolumn{1}{c|}{}                    & \multicolumn{1}{c|}{2}    & \multicolumn{1}{c|}{\textbf{16.81}} & \multicolumn{1}{c|}{\textbf{16.95}} & \multicolumn{1}{c|}{\textbf{134.71}} & \multicolumn{1}{c|}{\textbf{129.93}} & \multicolumn{1}{c|}{0.4568}          & \multicolumn{1}{c|}{0.4524}          & \multicolumn{1}{c|}{\textbf{0.6032}} & \multicolumn{1}{c|}{\textbf{0.6047}} & \multicolumn{1}{c|}{\textbf{0.1227}} & \textbf{0.1209} \\
\multicolumn{1}{c|}{}                    & \multicolumn{1}{c|}{4}    & \multicolumn{1}{c|}{17.18}          & \multicolumn{1}{c|}{17.33}          & \multicolumn{1}{c|}{126.82}          & \multicolumn{1}{c|}{124.19}          & \multicolumn{1}{c|}{\textbf{0.4547}} & \multicolumn{1}{c|}{\textbf{0.4516}} & \multicolumn{1}{c|}{0.5932}          & \multicolumn{1}{c|}{0.5939}          & \multicolumn{1}{c|}{0.1088}          & 0.1073          \\ \hline
\end{tabular}
\label{tab:ablation:alpha}
\end{table*}